\PassOptionsToPackage{table}{xcolor}
\documentclass[11pt]{article}
\usepackage[preprint]{acl}

\usepackage{times}
\usepackage{latexsym}

\usepackage[T1]{fontenc}
\usepackage[utf8]{inputenc}
\usepackage{microtype}
\usepackage{inconsolata}
\usepackage{graphicx}

\usepackage{url}
\usepackage{booktabs}
\usepackage{tabularx}
\usepackage{enumitem}
\usepackage{amsmath}
\usepackage{graphicx}
\usepackage{subcaption}
\usepackage{longtable}
\usepackage{pdflscape}
\usepackage{multirow} 
\usepackage{subcaption}
\usepackage{wrapfig}
\usepackage{lineno}
\usepackage[most]{tcolorbox}
\usepackage{array}
\usepackage{tabularx}
\usepackage[table]{xcolor}
\usepackage{makecell}
\usepackage{booktabs}
\usepackage{cleveref}
\usepackage{xurl}
\usepackage{seqsplit}
\usepackage{graphicx}
\usepackage{hyperref}

\usepackage{tikz}
\usetikzlibrary{arrows.meta, positioning}
\newcommand{\breaktt}[1]{\texttt{\seqsplit{#1}}}
\allowdisplaybreaks

\definecolor{GlobalCol}{RGB}{235,235,235}
\definecolor{DomainCol}{RGB}{222,235,247}
\definecolor{LLMCol}{RGB}{230,245,230}

\definecolor{ArxivCol}{RGB}{232,242,252}
\definecolor{RedditCol}{RGB}{255,239,230}
\definecolor{StoryCol}{RGB}{243,233,248}
\definecolor{WikiHowCol}{RGB}{232,247,232}
\definecolor{WikiCol}{RGB}{240,240,240}

\definecolor{datasetHeader}{HTML}{2F5E85}
\definecolor{humanRow}{HTML}{F3F3F3}
\definecolor{genRow}{HTML}{EEF6FB}
\definecolor{editRow}{HTML}{F8DDE8}

\definecolor{darkblue}{rgb}{0, 0, 0.5}
\hypersetup{colorlinks=true, citecolor=darkblue, linkcolor=darkblue, urlcolor=darkblue}

\newcommand{\dataset}{GEN}

\title{AI Writers Have a Consistent Stylometric Footprint, but AI Editors Do Not}

\author{First Author \\
  Affiliation / Address line 1 \\
  Affiliation / Address line 2 \\
  Affiliation / Address line 3 \\
  \texttt{email@domain} \\\And
  Second Author \\
  Affiliation / Address line 1 \\
  Affiliation / Address line 2 \\
  Affiliation / Address line 3 \\
  \texttt{email@domain} \\}

\author{
Zhengyang Shan\textsuperscript{1,*}, 
~~Yukyung Lee\textsuperscript{1,*}, 
~~Sophie Hao\textsuperscript{1}
\\ \textsuperscript{1}Boston University\\
\texttt{\{shanzy,ylee5,uu\}@bu.edu} \\ 
}

\author{
Zhengyang Shan\textsuperscript{1,*}, 
~~Yukyung Lee\textsuperscript{1,*}, 
~~Sophie Hao\textsuperscript{1}
\\ \textsuperscript{1}Boston University\\
\texttt{\{shanzy,ylee5,uu\}@bu.edu} \\ 
\\[1.2em]
\href{https://github.com/ZhengyangShan/stylometric-footprint}{%
\raisebox{-0.15em}{\includegraphics[height=1.05em]{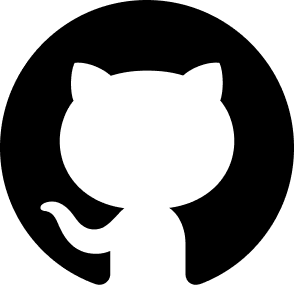}}%
\hspace{0.3em}\texttt{GitHub}}
\hspace{2em}
\href{https://huggingface.co/datasets/szyszy/GEN}{%
\raisebox{-0.15em}{\includegraphics[height=1.05em]{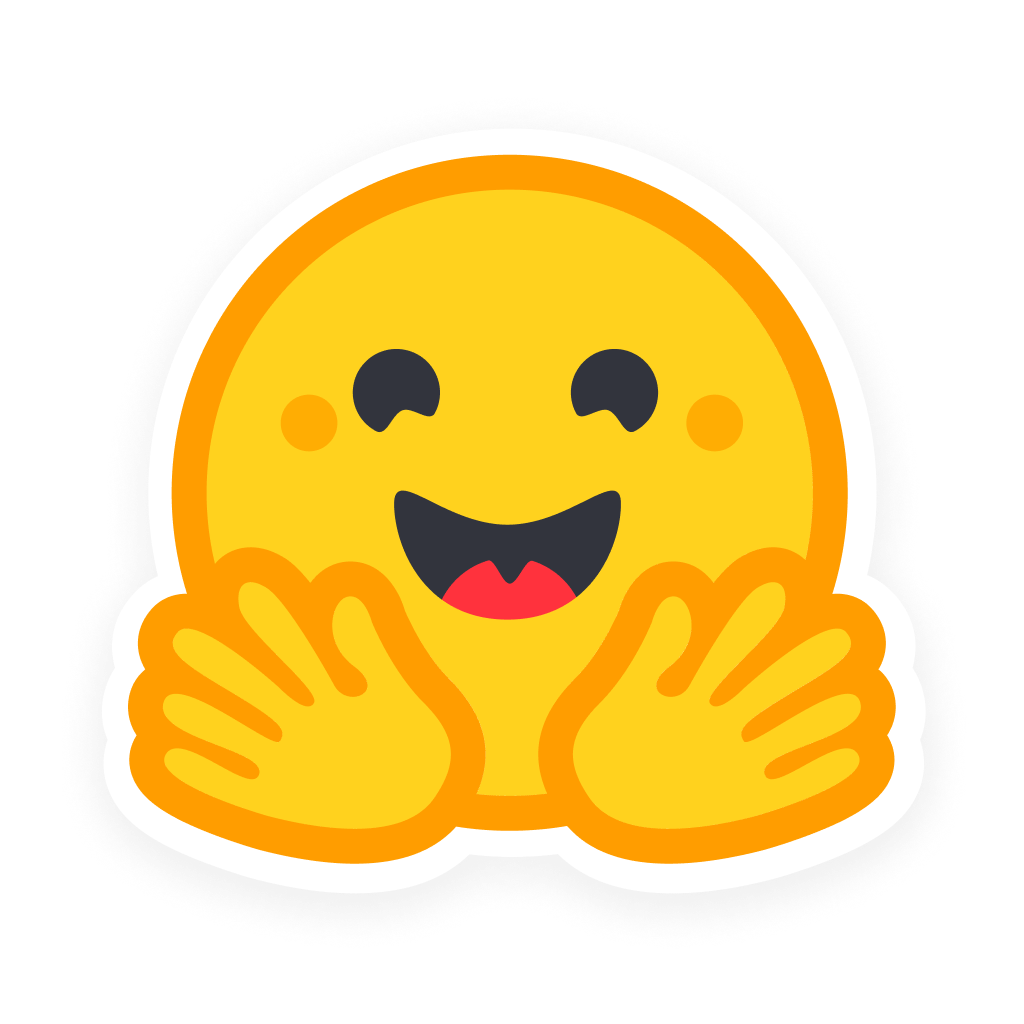}}%
\hspace{0.3em}\texttt{Hugging Face}}
}

\begin{document}
\maketitle
\begingroup
\renewcommand{\thefootnote}{}
\footnotetext{%
  \textsuperscript{\ensuremath{*}}Equal contribution.
}
\endgroup


\begin{abstract}
Text generated by large language models (LLMs) has been shown to be stylometrically distinct from human-written text \citep{andreDetectingAIAuthorship2023, shahDetectingUnmaskingAIGenerated2023, oparaStyloAIDistinguishingAIGenerated2024, soto2024fewshot, liLinguisticDifferencesAI2025, selviogluFeatureExtractionAnalysis2025}. But LLMs are increasingly used not only to generate text but also to edit human writing, and it is unclear whether the two leave the same trace. We show that AI generation leaves a consistent ``stylometric footprint'': a small subset of features, primarily entropy and lexical diversity, consistently separates AI-generated text from human writing across 8 LLMs and 5 domains, while the remaining features depend heavily on the domain and generator. AI editing, however, does not reproduce the same footprint. Relative to their human-written sources, AI-edited texts show only a small increase in lexical diversity and a decrease in entropy, rather than the joint increase that characterizes AI generation. Lexical density, which contributes little to generation, instead becomes the dominant editing-associated signal. Stylometric features therefore separate AI-edited text from AI-generated text but are substantially less effective at separating it from human-written text. Our results suggest that ``AI text'' is not a single phenomenon: generation and editing leave qualitatively different stylometric traces and should be studied separately. 
\end{abstract}

\begin{figure}[t]
    \centering
    \includegraphics[width=\linewidth]{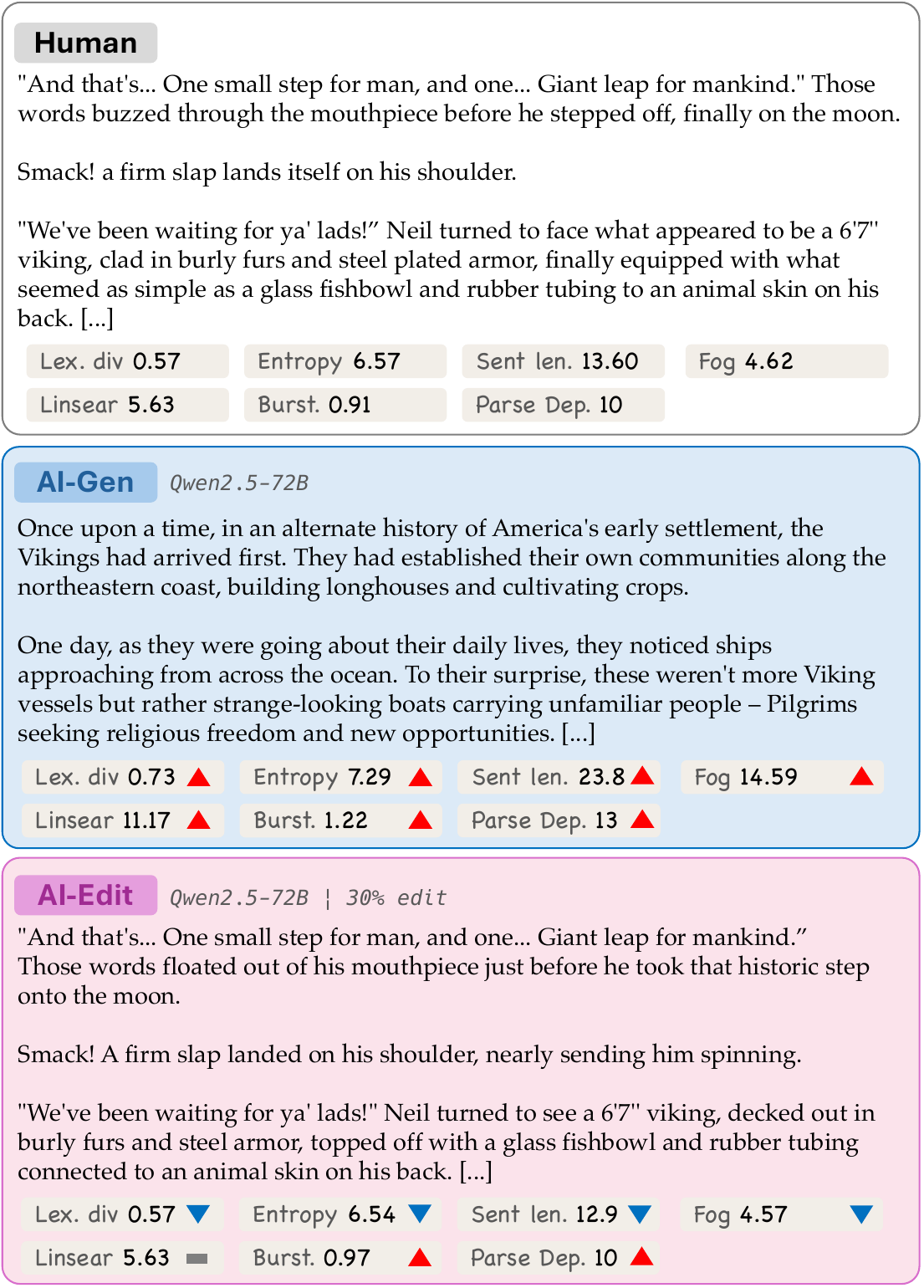}
    \caption{Overview of our pipeline. We compute 14 stylometric features across 5 domains and conduct two analyses: (1) which features reliably distinguish AI-generated text from human writing, and (2) whether AI editing shows the same patterns as generation.}
\label{fig:main}
\end{figure}

\section{Introduction}

AI assistants based on large language models (LLMs) are now a ubiquitous tool for writing \citep{zhao2026surveylargelanguagemodels}. Beyond generating full passages, LLMs are used to revise, rewrite,
paraphrase, and refine human-authored content \citep{chatterji2025people}. As AI-assisted writing becomes ever more sophisticated in its methods and diverse in its domains of application, it is becoming increasingly difficult to detect the use of AI in writing. In particular, there is reason to believe that the specific role played by an AI assistant in the writing process has ramifications for our ability to characterize AI-written text. For example, text that is fully generated by AI may very well exhibit different stylometric properties from human-written text that has been lightly edited using AI. 

Current methods for detection of AI-generated text range from zero-shot statistical tests \citep{mitchell2023detectgptzeroshotmachinegeneratedtext, bao2024fastdetectgpt, hans2024spotting} to fine-tuned neural classifiers \citep{guo2023closechatgpthumanexperts, bhattacharjee-etal-2023-conda, adam2026gptzerorobustdetectionllmgenerated}, and  feature-based approaches have been proposed in the interest of interpretability \citep{FrohlingZubiaga2021, shahDetectingUnmaskingAIGenerated2023, oparaStyloAIDistinguishingAIGenerated2024}. These methods rely on the assumption, tested in this paper, that text written with AI assistance can be characterized by a common set of statistical properties, which generalize across models, domains, and levels of AI involvement. If this assumption turns out to be incorrect, 
then detectors trained on fully AI-generated text may fail to transfer to AI-edited text despite strong in-distribution accuracy.

Concretely, we focus on the following questions:
\begin{itemize}[widest=RQ2., leftmargin=*]
    \item[\textbf{RQ1.}] Does AI-generated text have a consistent ``style''? 
    \item[\textbf{RQ2.}] Does AI-edited text adhere to this style?
\end{itemize} 
To answer these questions, we perform a stylometric analysis of 45,000 human-written and AI-generated texts across 8 LLMs and 5 writing domains, together with a large-scale corpus of 273,420 AI-edited texts generated using 3 LLMs across 31 editing prompts and 8 editing categories. We compare our human-written, AI-generated, and AI-edited texts along 14 interpretable features covering categories such as lexical diversity, entropy, readability, punctuation, sentence structure, and character composition.


Our results answer RQ1 in the affirmative. Across generators and domains, we find that AI-generated text has a ``stylometric footprint'' characterized by a small subset of features, primarily entropy and lexical diversity, that reliably distinguish it from human-written text. RQ2, however, is answered in the negative. Relative to their human-written sources, AI-edited texts show only a small increase in lexical diversity and a decrease in entropy, rather than the joint increase that characterizes fully AI-generated text. Lexical density instead emerges as the dominant effect of editing, despite contributing relatively little to the distinction between human-written and AI-generated text. Thus, editing does not simply move human writing toward a weaker version of the generation footprint.

AI editing does leave stylometric signals, but its footprint differs substantially from that of AI generation. Its detectability also varies by editing type, edit ratio, and model. We further find that stylometric and neural approaches capture complementary signals: stylometric models are substantially stronger at separating edited from generated text, while neural editing detectors are substantially more effective at identifying whether editing occurred at all. Combining the two substantially improves performance across both settings.

Overall, our results suggest that detectability is not a single property of ``AI text,'' but depends strongly on how LLMs are used. Rather than treating AI editing as a weaker form of AI generation, the evidence presented here suggests that generation and editing produce partially distinct stylometric signatures that should be studied separately.

We make the following contributions:
\begin{itemize}[leftmargin=*]
    \item We construct a benchmark dataset comprising 45,000 human and AI-generated texts across 8 LLMs and 5 domains, together with 273,420 AI-edited texts spanning 31 editing prompts and multiple editing categories.
    
    \item We show that a small subset of stylometric features, primarily entropy and lexical diversity, remains consistently informative across generators and domains, while most features are strongly condition-dependent.
    
    \item We provide a stylometric comparison of AI generation and AI editing at scale, showing that the two exhibit qualitatively different feature patterns, including reversals in the direction of several dominant generation-related signals.
    
    \item We demonstrate that stylometric and neural approaches capture complementary signals, and that combining them substantially improves AI-editing detection performance.
\end{itemize}

\section{Related Work}
\label{sec:related}

\subsection{AI-Generated Text Detection}
\label{sec:related:detection}

AI-generated text detection has developed mainly along two lines: zero-shot statistical methods and supervised classifiers. Zero-shot methods such as DetectGPT~\citep{mitchell2023detectgptzeroshotmachinegeneratedtext}, Fast-DetectGPT~\citep{bao2024fastdetectgpt}, and Binoculars~\citep{hans2024spotting} use likelihood, curvature, perplexity, or cross-perplexity to identify machine-generated text without task-specific training. They can perform well, but typically require reference language models and do not reveal which textual properties distinguish human-written from AI-generated text. Supervised methods, including fine-tuned RoBERTa models~\citep{wang2023implementingbertfinetunedroberta}, BiLSTM detectors~\citep{utku2025mila}, ensemble systems~\citep{Ensemble-IDS2025}, and GPTZero~\citep{adam2026gptzerorobustdetectionllmgenerated}, often achieve high in-distribution accuracy. Other work studies unseen-generator generalization through contrastive learning \citep{bhattacharjee-etal-2023-conda, zhang2024llmdetect} and fine-grained detection of LLM-produced text \citep{teja2025fine}. Related benchmarks capture complementary AI-text settings. M4 evaluates machine-generated text detection across generators and domains \citep{wang-etal-2024-m4}, while BEEMO studies expert- and LLM-edited machine-generated outputs \citep{artemova-etal-2025-beemo}. EditLens instead focuses on detecting and quantifying the extent of LLM editing in text \citep{thai2026editlens}. GEN is complementary to these resources: it places human-written, AI-generated, and LLM-edited human text in the same interpretable feature space, allowing us to directly test whether AI editing exhibits the same stylometric footprint as AI generation.


\subsection{Stylometric Features for Authorship Attribution and Detection}
\label{sec:related:stylometry}

Stylometric analysis has long studied measurable linguistic habits in authorship attribution, using features such as function word frequencies, vocabulary richness, character and POS n-grams, readability scores, and Burrows' Delta~\citep{stamatatos2009survey, koppel2009computational, burrows2002delta}. In AI-generated text detection, feature-based classifiers using lexical diversity, stop-word ratios, syntactic features, and readability measures can compete with heavier neural models~\citep{FrohlingZubiaga2021}, and stylometric features can complement pre-trained LMs~\citep{kumarage2023stylometric, oparaStyloAIDistinguishingAIGenerated2024}. Recent work has also explored learned representations of writing style for machine-generated text detection. \citet{soto2024fewshot} transfer style representations learned from human authorship to few-shot detection and attribution of machine-generated text. In contrast, we use a compact set of directly interpretable features to identify which specific stylistic properties remain stable across generators, domains, and modes of AI involvement. Related work has also evaluated whether LLMs can imitate individual writing styles using authorship attribution, authorship verification, stylometric metrics, and AI detection \citep{wang-etal-2025-catch}. Related work in automated essay scoring shows that prompt-independent surface features, including lexical diversity, vocabulary richness, and readability metrics, support strong cross-prompt generalization without complex neural architectures~\citep{li-ng-2024-conundrums, hou2025improve}. These results suggest that interpretable, content-insensitive features may support transfer. We build on this idea by evaluating which stylometric features remain stable across generators, domains, and AI-editing settings.

\section{\dataset: A Multi-Domain Corpus of AI-Edited Texts}
\label{subsec:dataset}


\begin{table}[h]
\centering
\footnotesize
\setlength{\tabcolsep}{3pt}
\renewcommand{\arraystretch}{1.3}
\setlength{\aboverulesep}{0pt}
\setlength{\belowrulesep}{0pt}

\begin{tabularx}{\columnwidth}{
>{\raggedright\arraybackslash}p{0.15\columnwidth}
>{\centering\arraybackslash}p{0.22\columnwidth}
>{\centering\arraybackslash}p{0.12\columnwidth}
>{\centering\arraybackslash}p{0.14\columnwidth}
>{\raggedleft\arraybackslash}X
}
\toprule
\rowcolor{datasetHeader}
\textcolor{white}{\textbf{Corpus}} &
\textcolor{white}{\textbf{Docs/domain}} &
\textcolor{white}{\textbf{LLMs}} &
\textcolor{white}{\textbf{Prompts}} &
\textcolor{white}{\textbf{Total}} \\
\midrule
\rowcolor{humanRow}
Human
& 1,000
& --
& --
& 4,998 \\
\rowcolor{genRow}
AI-Gen
& 1,000
& 8
& --
& 39,916 \\
\rowcolor{editRow}
AI-Edit
& 200 seeds
& 3
& 31
& 273,420 \\
\bottomrule
\end{tabularx}
\caption{
Analysis corpus across five domains: Wikipedia, WikiHow, Reddit, ArXiv, and Story Generation. For Human and AI-Gen, \textit{Docs/domain} denotes the number of documents sampled or generated per domain; for AI-Edit, it denotes the number of human-written seed documents edited per domain. \textit{Total} gives the number of documents retained in the final analysis corpus after filtering and train/test holdouts.}
\label{tab:dataset_structure}
\end{table}

\begin{figure}
    \centering
    \includegraphics[width=\columnwidth]{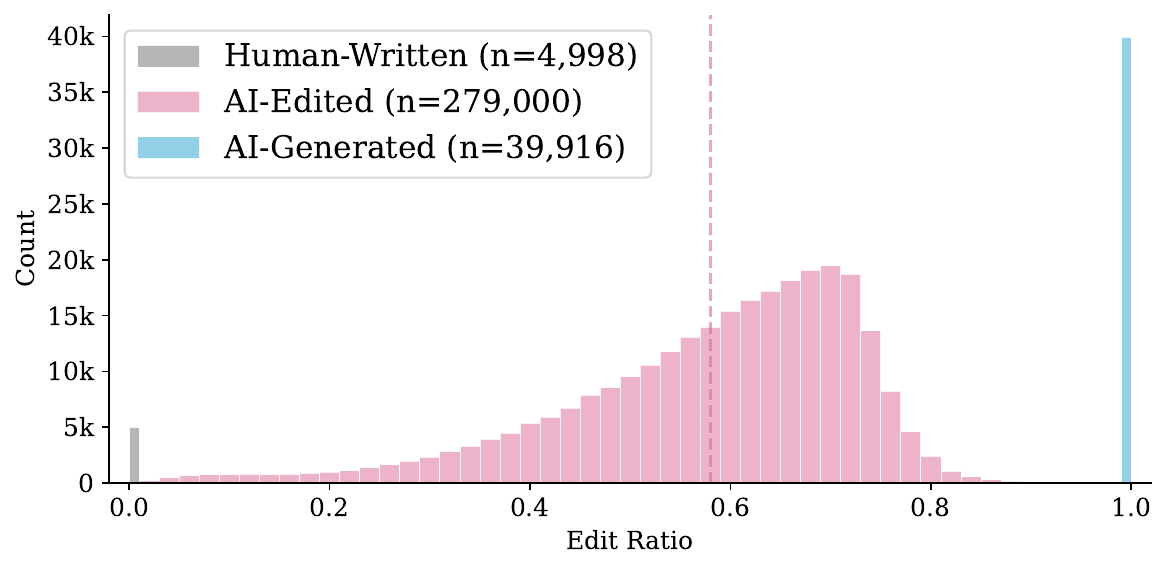}
    \caption{Edit ratio distribution in GEN.}
    \label{fig:GEN_histogram}
\end{figure}

This section introduces \dataset  (\textbf{G}enerated, \textbf{E}dited, and \textbf{N}aturalistic), a corpus of English-language text documents drawn from five \textit{domains}---Wikipedia, WikiHow, arXiv, Reddit QA, and Reddit Stories---each written with varying levels of AI involvement. We construct \dataset\ by augmenting human-written English documents from the M4 corpus \citep{wang-etal-2024-m4} and the \textsc{WritingPrompts} corpus \citep{fanHierarchicalNeuralStory2018} with AI-generated and AI-edited versions of these documents.  

\paragraph{Domains.} Human-written documents from the Wikipedia, WikiHow, arXiv, and Reddit QA domains are sampled from the corresponding domains of the M4 corpus,\footnote{\citet{wang-etal-2024-m4} refer to the Reddit QA domain as \textit{Reddit ELI5}.} while Reddit Stories documents are sampled from the \textsc{WritingPrompts} corpus. Wikipedia and WikiHow documents consist of full webpages, and arXiv documents consist of abstracts. Reddit QA documents are extracted from posts and comments in the subreddits \texttt{r/explainlikeimfive}, \texttt{r/AskScience}, and \texttt{r/AskHistorians}, while Reddit Stories documents are extracted from \texttt{r/WritingPrompts}.

\paragraph{Edit Ratio.}
We quantify the level of AI authorship for each document using its \textit{edit ratio}, defined as the word-level Levenshtein distance between a human-written document and its AI-edited version, normalized by the length of the longer document. We assign an edit ratio of 0 to human-written documents and an edit ratio of 1 to documents in the AI-generated condition. The resulting distribution of edit ratios in GEN is shown in \autoref{fig:GEN_histogram}.

\paragraph{AI Generation.} We use LLM prompts from \citet{wang-etal-2024-m4} to generate novel Wikipedia, WikiHow, arXiv, and Reddit QA documents. Reddit Stories documents are generated using prompts extracted from \breaktt{r/WritingPrompts} posts. 1,000 documents in each domain are generated using the following LLMs: \breaktt{Qwen-2.5-7B}, \breaktt{Qwen-2.5-72B}, \breaktt{Gemma-3-12B}, \breaktt{Gemma-3-27B}, \breaktt{Llama-3.1-8B}, \breaktt{Llama-3.3-70B}, \breaktt{GPT-oss-20B}, and \breaktt{GPT-oss-120B} \citep{grattafiori2024llama3herdmodels,qwen2025qwen25technicalreport,gemmateam2025gemma3technicalreport,openai2025gptoss120bgptoss20bmodel}.

\paragraph{AI Editing.} We apply AI editing to 200 human-written documents sampled from each domain (1,000 in total), using prompts from \citet{thai2026editlens}.\footnote{AI editing refers to LLM edits applied to fully human-written documents; we do not consider editing of AI-generated text.} Edited versions of each human-written document are generated using 31 unique prompts covering 8 edit types: grammar and mechanics, fluency and flow, clarity and precision, tone and style, paraphrasing, adding detail, structure and organization, and concision. Each prompt is applied with three target edit ratios (30\%, 50\%, 70\%), which specify the intended degree of editing but do not guarantee an exact amount of textual change. We therefore compute the actual edit ratio directly from each output using the word-level Levenshtein measure, and use this measured ratio in all analyses. The full list of prompts is provided in Appendix~\ref{app:editing_prompts}. Since editing requires greater instruction-following abilities than generation, only the three largest models for AI generation, \texttt{Qwen-2.5-72B}, \texttt{Llama-3.3-70B}, and \texttt{GPT-oss-120B}, are used for editing.\footnote{All text generation and editing experiments were conducted using vLLM v0.10.2~\citep{10.1145/3600006.3613165} with a maximum output length of 1,024 tokens, temperature 0.7, top-p 0.9, and random seed 42. We used NVIDIA A100 80GB, A6000 48GB GPUs.} 279,000 edited texts are generated in total; 273,420 of these are used for analysis, after holding out 20 seed documents for classifier training (Appendix~\ref{app:implementation-detail}).

\section{Experimental Setup: Characterizing Stylometric Footprints}
\label{sec:method}

The goal of this paper is to determine whether AI-generated text can be consistently and parsimoniously described by a ``stylometric footprint,'' and whether AI-edited text adheres to this footprint. We operationalize these ideas by conducting experiments (Sections~\ref{sec:aigen_analysis} and \ref{sec:phase2-results}) where linear classifiers are trained on subsets of \dataset, represented using only stylometric features, and analyzed using metrics of feature importance and robustness across experimental conditions.


\subsection{Stylometric Features}

\begin{table}
    \centering
    \footnotesize
    {
    \renewcommand\tabularxcolumn[1]{m{#1}}
    \begin{tabularx}{\linewidth}{>{\raggedright\arraybackslash}m{1.75cm} X}
        \toprule
        \textbf{Category} & \textbf{Features} \\\midrule
        Lexical & lexical diversity, lexical density, \% long words, document length (words), document length (sentences), mean sentence length (words) \\\midrule
        Orthographic & \% vowels, \% consonants, \% punctuation \\\midrule
        Syntactic &  max parse tree depth \\\midrule
        Information-Theoretic & entropy, burstiness \\\midrule
        Readability & Gunning Fog, Linsear Write \\\bottomrule
    \end{tabularx}
    }
    \caption{Stylometric features considered in our study. See \autoref{tab:features} for full definitions and formulas and \autoref{app:features} for selection criteria.}
    \label{tab:features-summary}
\end{table}
Our study considers 14 stylometric features, shown in \autoref{tab:features-summary}, spanning five categories: \textit{lexical}, \textit{orthographic}, \textit{syntactic}, \textit{information-theoretic}, and \textit{readability}. 
Our lexical, orthographic, and syntactic features are straightforward to interpret. Information-theoretic features capture properties of the word-frequency distribution: entropy~\citep{shannon1948} measures distributional uniformity, while burstiness measures frequency dispersion. Our readability features, Gunning Fog~\citep{gunning1952} and Linsear Write, estimate the level of education required to read the passage.

\subsection{Linear Classifiers}
\label{sec:logistic-regression}

Our linear classifiers are trained using logistic regression. Each classifier outputs the predicted probability that a text belongs to the \textit{positive class}:
\[
\hat{y} = \sigma\!\left(\beta_0 + \sum_{i=1}^{14} \beta_i x_i\right)
\]
where $\sigma(\cdot)$ is the logistic (sigmoid) function, $x_i$ are the normalized stylometric features, and $\beta_i$ the learned coefficients. The positive class is defined per task. All classifiers use $\ell_2$ regularization ($C=1.0$), balanced class weights, and the \texttt{liblinear} solver.
Each classifier is trained and evaluated over ten independent runs. In each run, we sample a class-balanced subset of \dataset, and then split data into 80\% train and 20\% test using stratified sampling. All reported accuracies are means over the ten held-out test sets; standard deviations across runs are reported to characterize result stability. 
Features are standardized using the training data before evaluation on the test set.

\subsection{Feature Importance Metrics}
\label{subsec:phase1}

We quantify the importance of each feature $x_i$ using two complementary metrics. \textit{Coefficient importance} is defined as the mean value of $|\beta_i|$ across the ten runs for each classifier and reflects how strongly a standardized feature influences the classifier's log-odds, conditional on the other features. \textit{Permutation importance} is defined as the drop in held-out accuracy attained when the classifier is retrained with the values for $x_i$ shuffled in the training dataset. Whereas coefficient importance measures how strongly the fitted model uses a feature, permutation importance more directly measures that feature's contribution to predictive performance.
When estimating permutation importance, 10 shuffle-and-retrain runs are performed for each of the ten runs of each classifier. 
\section{The Stylometric Footprint of AI-Generated Text}
\label{sec:aigen_analysis}

\begin{figure}
    \centering
    \includegraphics[width=\columnwidth]{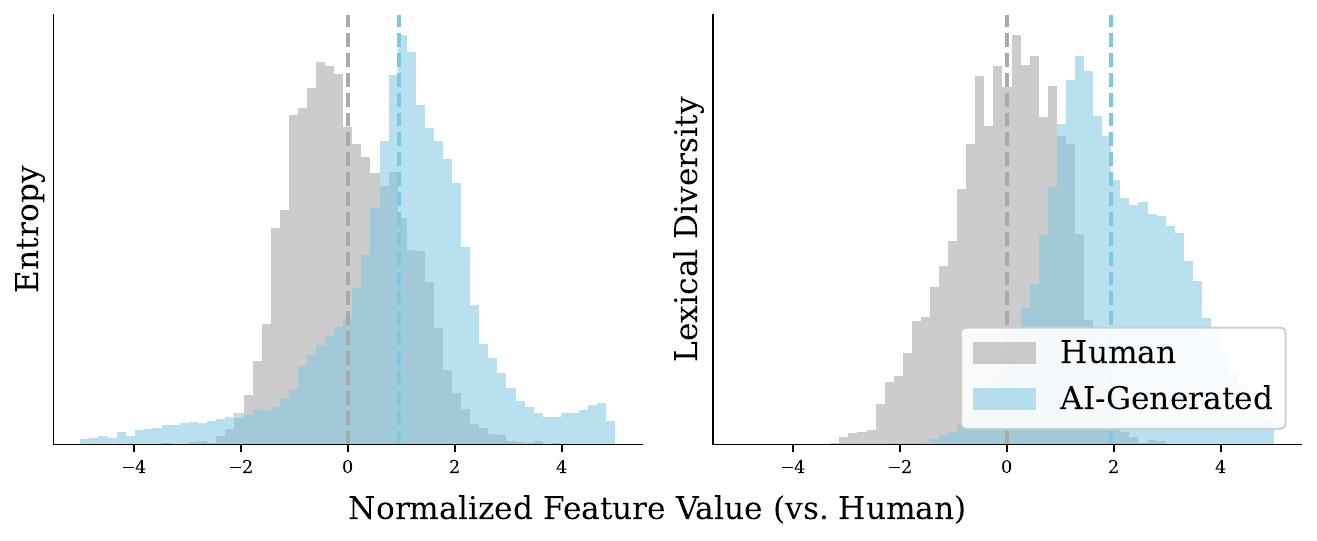}
    \caption{Distribution of entropy and lexical diversity, for human-written and AI-generated documents. Feature values are normalized to the human distribution ($z$-score units). Dashed lines indicate group means.}
    \label{fig:feature_shift}
\end{figure}

In this section, we show that AI-generated text is characterized by high entropy and high lexical diversity compared to human-written text. As illustrated in \autoref{fig:feature_shift}, the mean entropy of the AI-generated documents in \dataset\ is a full standard deviation higher than that of the human-written documents, and the lexical diversity of our AI-generated documents is two standard deviations above that of our human-written documents. These properties form a ``stylometric footprint'' that distinguishes AI-generated text from human-written text. To show this, we conduct an experiment based on the following axioms.
\begin{itemize}
    \item \textbf{Existence:} If AI-generated text has a ``stylometric footprint,'' then a linear classifier should be able to distinguish AI-generated text from human-written text using only stylometric features.
    \item \textbf{Parsimony:} If this footprint is \textit{parsimonious}, then the classifier's behavior should be driven by a small number of ``important'' features.
    \item \textbf{Consistency:} If this footprint is \textit{consistent}, then the same set of important features should drive the classifier's behavior, regardless of which domains and LLM generators are included in the training and testing data.
\end{itemize}
Our experiment aims to demonstrate that the stylometric footprint of AI-generated text is parsimonious and consistent.\footnote{Detailed experimental results are reported in \autoref{app:analysis_setting1}.}

\begin{figure*}[t]
    \centering    \includegraphics[width=\linewidth]{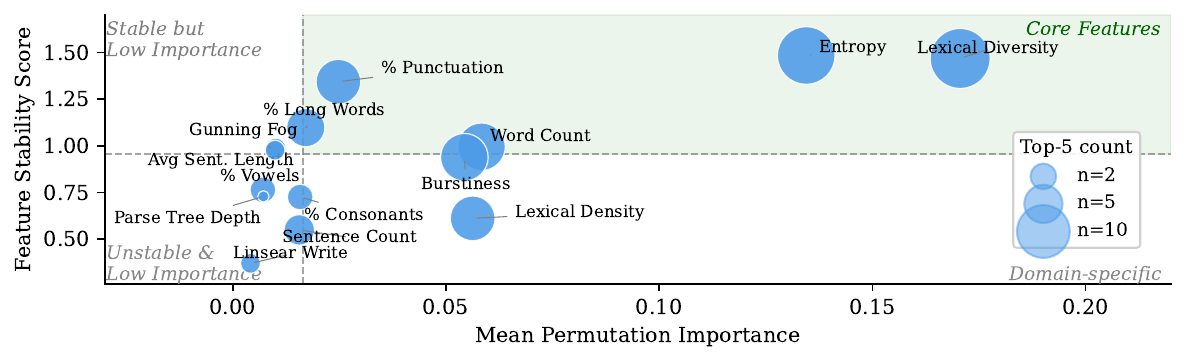}
    \caption{
    The \textit{feature stability score} ($y$-axis) measures whether a feature is consistently important for classification of AI-generated text across the 14 conditions in \autoref{tab:acc_summary} (Global, Per-Domain, Per-LLM). Higher scores indicate higher mean ($x$-axis) and lower standard deviation of permutation importance across conditions. \textit{Top-5 count} (bubble size) is the number of conditions where a feature ranks in the top five by permutation importance.
    }
    \label{fig:feature_robustness}
\end{figure*}

\subsection{Existence of a Stylometric Footprint}
\label{sec:stylometric_separation}

\begin{table}[t]
\centering
\footnotesize
\setlength{\tabcolsep}{4pt}
\renewcommand{\arraystretch}{1.5}
\begin{tabularx}{\columnwidth}{lXc}
\toprule
\textbf{Level} & \textbf{Condition} & \textbf{Acc.\ (\%)} \\
\midrule
\textit{Random} & --- & 50.00 \\
\midrule
\cellcolor{GlobalCol}Global & All & 91.74 \scriptsize{$\pm$0.61} \\
\midrule
\cellcolor{DomainCol}Per-Domain & ArXiv & 92.57 \scriptsize{$\pm$1.29} \\
\cellcolor{DomainCol} & Reddit QA & 92.67 \scriptsize{$\pm$1.21} \\
\cellcolor{DomainCol} & Reddit Stories& 95.83 \scriptsize{$\pm$0.99} \\
\cellcolor{DomainCol} & WikiHow & 95.72 \scriptsize{$\pm$1.01} \\
\cellcolor{DomainCol} & Wikipedia & 92.99 \scriptsize{$\pm$1.31} \\
\midrule
\cellcolor{LLMCol}Per-LLM & Gemma 12B & 98.15 \scriptsize{$\pm$0.32} \\
\cellcolor{LLMCol} & Gemma 27B & 98.11 \scriptsize{$\pm$0.27} \\
\cellcolor{LLMCol} & GPT-OSS 20B & 98.77 \scriptsize{$\pm$0.25} \\
\cellcolor{LLMCol} & GPT-OSS 120B & 79.45 \scriptsize{$\pm$0.87} \\
\cellcolor{LLMCol} & Llama 8B & 97.78 \scriptsize{$\pm$0.31} \\
\cellcolor{LLMCol} & Llama 70B & 95.38 \scriptsize{$\pm$0.44} \\
\cellcolor{LLMCol} & Qwen 7B & 99.09 \scriptsize{$\pm$0.21} \\
\cellcolor{LLMCol} & Qwen 72B & 92.47 \scriptsize{$\pm$0.60} \\
\bottomrule
\end{tabularx}
\caption{Accuracy for classification of human-written vs.\ AI-generated text for subsets of \dataset\ (mean $\pm$ std over ten class-balanced runs). The \textit{Per-Domain} and \textit{Per-LLM} conditions only use documents from the same domain or LLM generator, respectively. The \textit{Global} condition uses documents from all domains and LLM generators.} 
\label{tab:acc_summary}
\end{table}

To establish the existence of a stylometric footprint, we show in \autoref{tab:acc_summary} that human-written text can be distinguished from AI-generated text using a linear classifier with only stylometric features. Accuracies of over 90\% are obtained whether texts from only one domain are used (\textit{Per-Domain} conditions) or whether texts from all domains are used (\textit{Global} condition). The same holds when texts from only one LLM generator are used (\textit{Per-LLM} conditions), except when the AI-generated texts were generated using GPT-OSS 120B. In this condition, mean accuracy is only 79.45, indicating that texts generated by this model are exceptionally human-like in terms of stylometrics.

\subsection{Parsimony of the Stylometric Footprint}

\begin{figure*}[t]
    \centering
    \includegraphics[width=\linewidth]{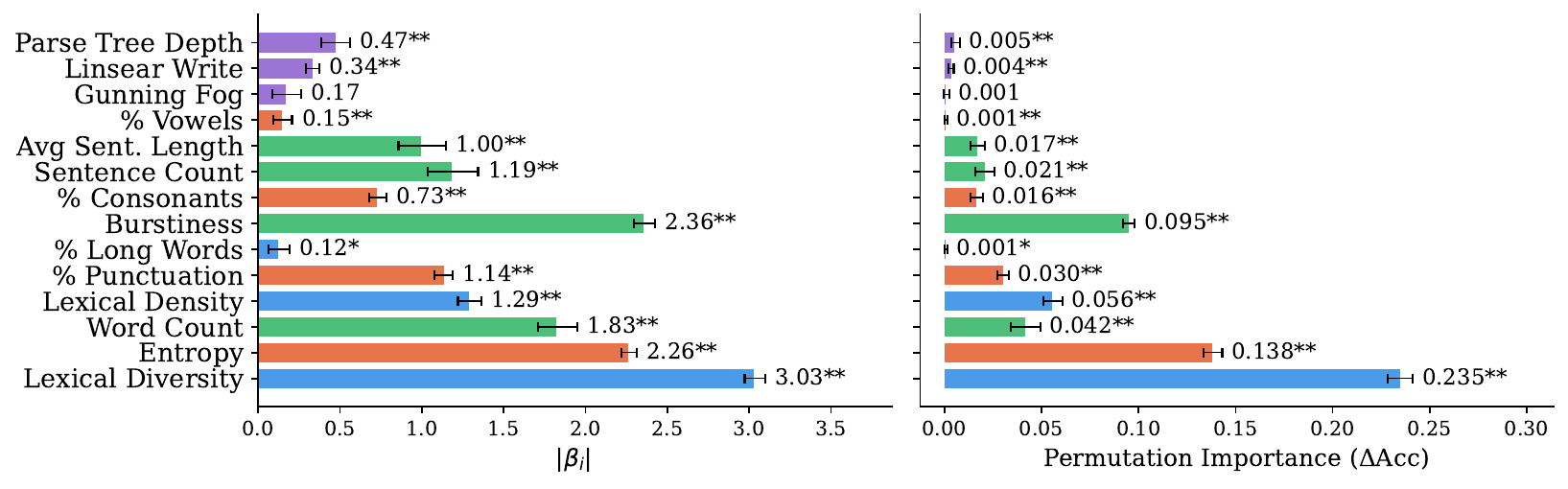}
    \caption{The coefficient importance ($|\beta_i|$) and permutation importance ($\Delta$Acc) of each feature in our linear classifiers for human-written vs.\ AI-generated text. Values are averaged across the ten training and testing runs from the Global condition (\autoref{tab:acc_summary}). Error bars indicate 95\% bootstrap confidence intervals; asterisks denote permutation-test significance (* $p<0.05$, ** $p<0.01$).}
    \label{fig:global_importance}
\end{figure*}

\autoref{fig:global_importance} shows feature importance scores for the classifier from the Global condition, which uses texts from all domains and LLM generators. Lexical diversity, entropy, and burstiness are the three most important features for this classifier under both metrics. Under permutation importance, each of these features produces a substantially larger held-out accuracy drop than most of the remaining features, with lexical diversity and entropy contributing the strongest individual effects. 
While several other features have relatively large coefficients, their substantially lower permutation importance indicates that strong
influence on the fitted log-odds does not necessarily translate into a
large contribution to held-out classification performance. 
These results therefore indicate that the three most important features for our model form a parsimonious characterization of AI-generated text. 

\subsection{Consistency of the Stylometric Footprint}
\label{sec:cross_condition_variability}
\label{sec:feature_rank_stability}

Finally, we show that the stylometric footprint of AI-generated text is consistent across domains and LLM generators. To determine which features are consistently important for detecting AI-generated text, we define the \textit{feature stability score} of a feature $x_i$ to be $s_i = \mu_i/\sigma_i$, where $\mu_i$ and $\sigma_i$ are the mean and standard deviation, respectively, of the permutation importance of $x_i$ measured across the 14 conditions shown in \autoref{tab:acc_summary} (Global, Per-Domain, and Per-LLM). In order to attain a high feature stability score, $x_i$ must be important for classification in all conditions, and its importance must exhibit little variation across conditions.

\autoref{fig:feature_robustness} shows that entropy and lexical diversity have the highest feature stability score, and their mean permutation importance across conditions is significantly higher than that of all other features. Lexical diversity appears among the five most important features for 13 of 14 conditions, and entropy for 12 of 14, both with substantially lower variance than other features. Although burstiness had the third highest permutation importance in the Global condition (\autoref{fig:global_importance}), this feature does not remain consistently important across domains and LLM generators. 
Burstiness, lexical density, word count, and \% punctuation, are moderately important across conditions but less stable than entropy or lexical diversity, and the remaining features are neither important nor stable.

\subsection{Robustness to Generation Settings}
\label{sec:generation-robustness}

\paragraph{Sampling temperature.}
To test whether the identified footprint is specific to a single decoding temperature, we regenerate 1,000 documents using Llama-3.1-8B at temperatures 0.3, 0.7, and 1.0, using 200 prompts from each domain and holding the prompts, remaining decoding parameters, and feature pipeline fixed. Classification accuracy changes only modestly across temperatures, from 82.2\% at temperature 0.3 to 80.7\% at 0.7 and 79.5\% at 1.0. Feature rankings also remain strongly correlated across temperature pairs (mean Spearman $\rho=0.89$, range $0.86$--$0.93$). Lexical diversity ranks first at all three temperatures, while entropy remains among the five most informative features throughout. Thus, although the relative importance of secondary features changes with temperature, the central footprint characterized by entropy and lexical diversity is not an artifact of the sampling temperature used in the main experiments. Full results are reported in Appendix~\ref{sec:temperature-ablation}.

\paragraph{Proprietary generators.}
We additionally evaluate 2,500 documents from each of two proprietary models, Gemini-3-Flash and GPT-5.4-mini, using 500 prompts from each of the five domains. The same stylometric classification setup achieves 82.1\% accuracy for Gemini-3-Flash and 78.6\% for GPT-5.4-mini. Lexical diversity remains informative across both models, ranking first for Gemini-3-Flash and fourth for GPT-5.4-mini. However, the secondary signals differ: character-level and readability features contribute more strongly for GPT-5.4-mini. These results extend the overall detectability pattern to proprietary generators while reinforcing that the precise feature ranking remains generator-dependent. Full results are provided in Appendix~\ref{sec:proprietary-generation}.
\begin{figure*}[t]
    \centering
    \includegraphics[width=\linewidth]{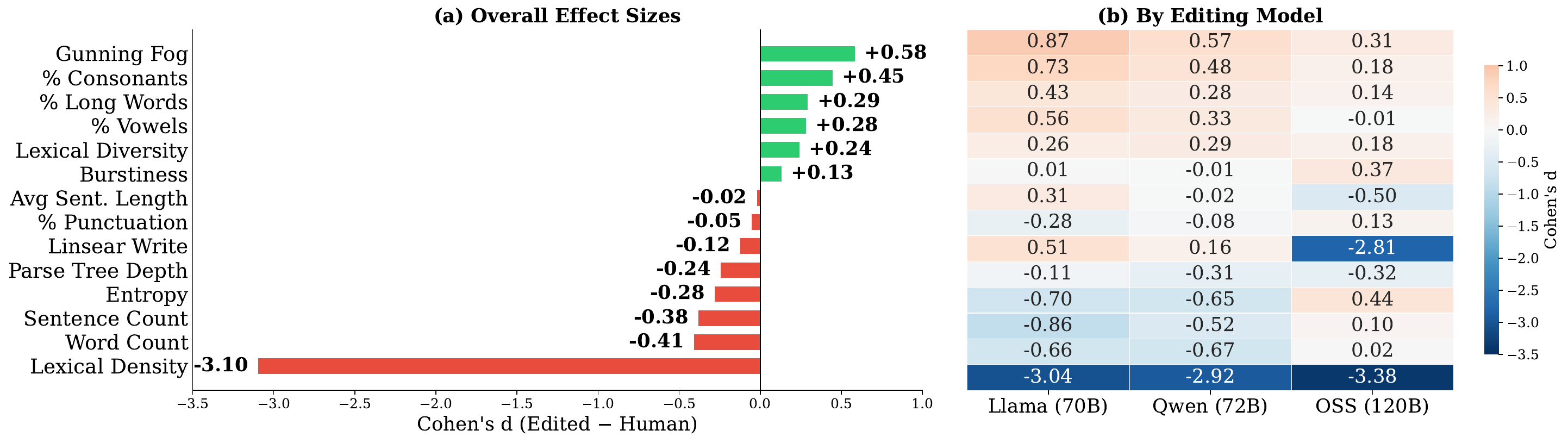}
    \caption{The effect size of AI editing on the stylometric properties of human-written documents, measured by Cohen's $d$. Positive values indicate that editing increases the feature; negative values indicate a decrease. (a) shows effect sizes estimated from all AI-edited documents. (b) shows effect sizes broken down by editing model.}
\label{fig:cohens_d_all}
\end{figure*}

\section{AI-Edited vs.\ AI-Generated Text}
\label{sec:phase2-results}

We have now shown in \autoref{sec:aigen_analysis} that fully AI-generated text has a parsimonious and consistent stylometric footprint, characterized by high entropy and high lexical diversity. In this section, we turn to text that LLMs have edited from human-written sources. We analyze how AI editing affects stylometric features and whether the patterns found in generation hold under editing. 

\subsection{Stylometric Effects of AI Editing}
\label{sec:editing_features}

\autoref{fig:cohens_d_all}(a) shows the effect size of AI editing on each stylometric feature of our human-written texts, measured using Cohen's $d$. Recall that the AI-edited documents of GEN are edited versions of the human-written documents in GEN. For each feature $x_i$, Cohen's $d$ is computed as follows:
\[
d_i = \frac{\mu\left(x_i^{\text{edited}} - x_i^{\text{human}}\right)}
{\sigma\left(x_i^{\text{edited}} - x_i^{\text{human}}\right)},
\]
where $x_i^{\text{edited}}$ and $x_i^{\text{human}}$ denote the value of $x_i$ for an AI-edited document and its human-written counterpart, respectively. The mean $\mu$ and standard deviation $\sigma$ are computed across all document pairs.

While AI-generated text is characterized by high lexical diversity and high entropy, \autoref{fig:cohens_d_all}(a) shows that, relative to the human-written source documents, AI editing increases lexical diversity only marginally ($d=+0.24$), while actually decreasing entropy ($d=-0.28$). Therefore, AI editing does not make human-written text stylometrically more similar to AI-generated text. Instead, the feature most impacted by AI editing is lexical density ($d=-3.10$): AI editing reduces the proportion of content words in a document, relative to function words. This feature, with a permutation importance of only 0.056 in \autoref{fig:global_importance}, is not a strong indicator of AI-generated text, even though it is the strongest stylometric indicator of AI editing.

\autoref{fig:cohens_d_all}(b) shows that editing by GPT-OSS 120B produces different stylometric effects compared to the other two AI editors used in \dataset, Llama 70B and Qwen 72B. Unlike the other models, GPT-OSS increases entropy ($d = +0.44$), keeps word count roughly constant ($d = +0.02$), and sharply decreases Linsear Write ($d = -2.81$).

\subsection{How Editing Differs from Generation}

The effect sizes in Section~\ref{sec:editing_features} show that editing does not move human-written text consistently toward the generation footprint. We next ask whether these differences persist when the stylometric features are considered jointly: do AI-edited texts occupy a distinct region of the feature space from AI-generated texts, and how close do they remain to their human-written sources? To answer this, we train a logistic regression classifier with the same setup as in Section~\ref{sec:logistic-regression} on three pairings: Human vs.\ AI-gen, AI-edit vs.\ AI-gen, and Human vs.\ AI-edit. For the first pairing, we apply the global generation classifier from \autoref{sec:aigen_analysis} to edited text in a zero-shot setting.


\begin{table}[t]
\centering
\setlength{\tabcolsep}{4pt}
\renewcommand{\arraystretch}{1.3}
\resizebox{\columnwidth}{!}{%
\begin{tabular}{lccc}
\toprule
Method & \makecell{Human vs.\\ AI-gen} & \makecell{AI-edit vs.\\ AI-gen} & \makecell{Human vs.\\ AI-edit} \\
\midrule
LR        & 0.97 & 0.98 & 0.80 \\
EL-R      & 0.97 & 0.71 & 0.94 \\
EL-L      & 1.00 & 0.85 & 0.97 \\
\midrule
LR$+$EL-R      & 0.99 & 0.97 & 0.91 \\
LR$+$EL-L      & 1.00 & 0.99 & 0.94 \\
EL-R$+$EL-L    & 1.00 & 0.80 & 0.98 \\
\midrule
LR$+$EL-R$+$EL-L & 1.00 & 0.96 & 0.96 \\
\bottomrule
\end{tabular}%
}
\caption{AUC across methods. LR denotes our stylometric logistic regression classifier, and EL-R and EL-L denote EditLens with \texttt{RoBERTa-large} and \texttt{Llama-3.2-3B} backbones, respectively. Ensemble models combine methods by averaging their scores.}
\label{tab:auc}
\end{table}

We compare against EditLens~\citep{thai2026editlens}, a state-of-the-art editing-aware model that operates on raw text rather than stylometric features. EditLens scores text from 0 to 1; we use both its \texttt{RoBERTa-large} (EL-R) and \texttt{Llama-3.2-3B} (EL-L) backbones, with thresholds set by grid search on the test set to maximize macro F1 (\Cref{app:implementation-detail}).

\begin{figure}[t]
\centering
\includegraphics[width=\columnwidth]{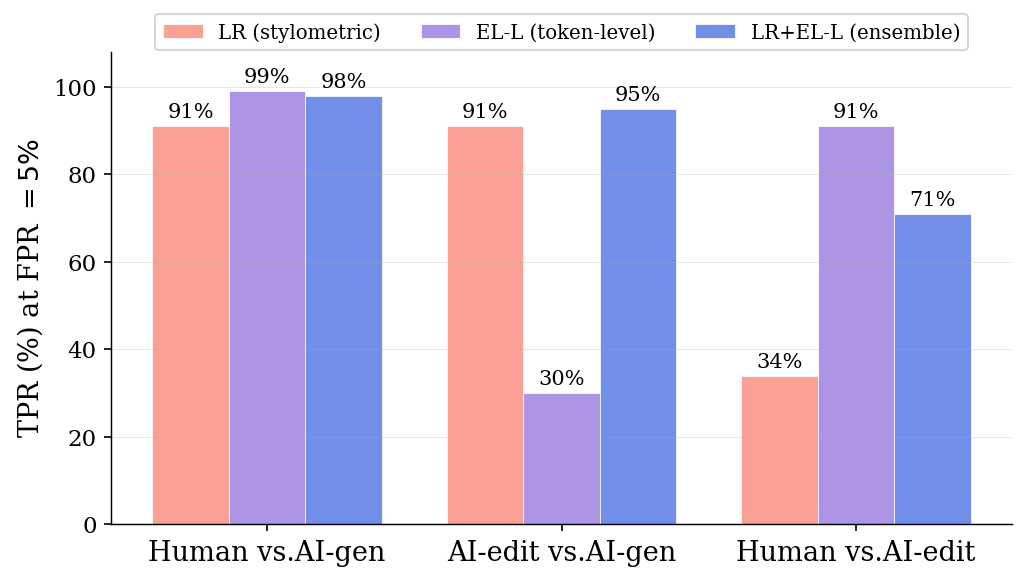}
\caption{TPR at 5\% FPR for LR, EL-L, and their ensemble across tasks.}
\label{fig:complementary}
\end{figure}

We report AUC (Table~\ref{tab:auc}) and TPR at FPR $=5\%$ (Figure~\ref{fig:complementary}) as measures of how separable each pair of classes is. The two methods are strong on different tasks. The stylometric features separate edited from generated text (AI-edit vs.\ AI-gen, AUC $0.98$), confirming at the classifier level that editing does not reproduce the generation footprint. But they are substantially less effective at separating edited text from human-written text (Human vs.\ AI-edit, AUC $0.80$): along these features, edited text stays close to the human original. EditLens shows the reverse pattern. It separates edited text from human writing (Human vs.\ AI-edit, AUC $0.97$, TPR $91\%$ for EL-L), capturing signals that are not represented by our 14 stylometric features, but is weaker at distinguishing edited from generated text. This ordering is unchanged under tuned thresholds (Appendix~\ref{app:threshold}), and separation increases with edit ratio.

The two signals are thus complementary, each strongest where the other is weakest: stylometric features on AI-edit vs.\ AI-gen (LR $0.98$ vs.\ EL-L $0.85$), EditLens on Human vs.\ AI-edit (EL-L $0.97$ vs.\ LR $0.80$). Combining them recovers the stronger signal on both (LR$+$EL-L, AUC $0.99$ and $0.94$; full breakdown in Appendix~\ref{app:full_results}). The takeaway is consistent across both: editing is stylometrically far from generation but close to human writing, the opposite of what the generation footprint would predict. A three-class classification over Human, AI-edit, and AI-gen shows the same pattern: AI-edited text is misclassified far more often as Human than as AI-gen, indicating that it is not a simple midpoint between the two (See Appendix~\ref{app:three_class}).

\section{Conclusion}

In this work, we presented a systematic stylometric comparison of AI-generated and AI-edited text. Our analysis shows that AI generation leaves a parsimonious and consistent footprint: among the 14 features considered, entropy and lexical diversity remain consistently informative across generators and domains, while most other features are strongly condition-dependent. Entropy and lexical diversity stay among the most important features under length controls and when individual generators or domains are held out, indicating that their importance does not depend on any single condition. The contrast between a small stable core and many condition-dependent features helps explain why stylometric signals may fail to transfer across settings.  AI editing, however, does not consistently reproduce this footprint. Relative to human-written sources, lexical diversity increases only weakly, while entropy moves in the opposite direction from the generation-associated signal. Lexical density instead becomes the dominant effect of editing, despite contributing comparatively little to generation detection. Stylometrically, edited text therefore remains closer to its human-written source than to fully generated text. These results show that generation and editing leave qualitatively different traces and should not be treated as a single phenomenon of ``AI text.'' We hope this work contributes to a better understanding of the different ways AI shapes written text.

\section*{Limitations}
\label{sec:limitations}

Several limitations should be considered when interpreting the results of this study.

First, the analysis focuses on a relatively small set of stylometric features designed to remain interpretable and computationally lightweight. Although these features capture meaningful cross-condition patterns, they do not represent the full space of possible linguistic or semantic signals. 

Second, the evaluated generators and domains, while diverse, do not exhaust the range of modern text-generation settings. Our main benchmark includes eight open-weight instruction-tuned LLMs across five domains, supplemented by a targeted analysis of two proprietary generators, Gemini-3-Flash and GPT-5.4-mini. Although the proprietary model results broaden our model coverage, this analysis is smaller in scale than the main benchmark, and our large-scale AI-editing experiments remain limited to three open-weight editors. Future models may exhibit different stylistic behaviors as generation systems continue to improve in controllability and human-like variation. Similarly, the domains considered here primarily involve English long-form text and may not generalize to multilingual, conversational, or highly specialized writing settings.

Third, the feature analyses remain associational rather than causal. High permutation importance or coefficient stability indicates that a feature is predictive under the evaluated conditions, but does not establish that the feature is intrinsically responsible for the observed separation between AI-generated and human-written text. Some features may correlate with broader stylistic or dataset-specific properties not directly measured in this work.


\bibliography{custom}

\appendix

\newpage

\appendix
\section{Human Text Collection}
\label{app:dataset}
We apply domain-specific sampling procedures to control for known confounds in each source. Example prompts are shown in Table~\ref{tab:prompt_examples}.

\paragraph{Wikipedia} articles in M4 dataset vary substantially in length, which affects the values of length-sensitive features (lexical diversity, number of words, entropy). To prevent length from becoming a confound, we stratify by prompt length using a median split at 107 characters, yielding two equal-sized bins (short: $\le$ 107 chars; long: $>$ 107 chars). We sample 500 texts from each bin for a total of 1,000. This stratification ensures that the Wikipedia sample is length-balanced and the human distribution of length-sensitive features does not artificially differ from AI-generated Wikipedia text due to sampling bias.

\paragraph{WikiHow} prompts in M4 range from very short titles (a few words) to long multi-sentence descriptions. We exclude prompts longer than 2,000 characters to remove outlier entries that may represent unusually complex items. After filtering, we sample 1,000 texts uniformly at random. 

\paragraph{ArXiv} abstracts in M4 are relatively homogeneous in length and structure. We sample 1,000 texts uniformly at random with no additional filtering.

\paragraph{Reddit} posts in M4 are associated with prompts that specify a writing voice (e.g., "answer in an expert confident voice", "answer in a casual voice"). Because voice style influences surface linguistic features, uniform random sampling could produce a voice-skewed human distribution. We apply a stratified sampling procedure to balance voice styles. Posts with the least frequent style ("expert confident", n=231) are included in full. For each remaining voice style, we sample an equal proportional share of the remaining 769 posts, distributed equally across the three other styles.

\paragraph{Story Generation} WritingPrompts dataset (euclaise/writingprompts, HuggingFace), a large collection of Reddit r/WritingPrompts posts paired with top-voted human-authored story responses. The dataset contains multiple prompt categories; we restrict to Writing Prompt (WP) entries only, identified by the prefix ``[ WP ]''. This tag denotes open-ended creative fiction prompts, as opposed to other prompt categories (Established Universe, Reality Fiction, etc.) that impose additional constraints on the story type. 
We further filter to prompts with at least 70 characters to exclude very short or underdeveloped prompts that may not provide sufficient context for story generation. After filtering, we sample 1,000 prompts uniformly at random.

\begin{table*}[t]
\centering
\small
\renewcommand{\arraystretch}{1.25}

\begin{tabularx}{\textwidth}{@{} p{2.0cm} X @{}}
\toprule
\textbf{Domain} & \textbf{Example Prompt} \\
\midrule

\textbf{Wikipedia} &
\begin{tcolorbox}[
    colback=gray!4,
    colframe=gray!45,
    boxrule=0.5pt,
    arc=2mm,
    left=1.5mm,right=1.5mm,
    top=1mm,bottom=1mm,
    width=\linewidth,
    before skip=0pt,
    after skip=0pt
]
Write a Wikipedia article with the title \textit{``Nevadaplano''}. The article should contain at least 250 words.
\end{tcolorbox}
\\

\midrule

\textbf{WikiHow} &
\begin{tcolorbox}[
    colback=blue!2,
    colframe=blue!45,
    boxrule=0.5pt,
    arc=2mm,
    left=1.5mm,right=1.5mm,
    top=1mm,bottom=1mm,
    width=\linewidth,
    before skip=0pt,
    after skip=0pt
]
Write a WikiHow article given a title and a headline using approximately 300 words.

\smallskip
\textbf{Title:}
\textit{``How to Make a Butterfly Origami''}

\smallskip
\textbf{Headline:}
\textit{``Start with a square piece of paper. Make a horizontal valley crease. Make a vertical valley crease through the center. Rotate the paper 45 degrees. \ldots{} Give your butterfly as a gift, or use as a decoration.''}

\smallskip
\textbf{Article content:}
\end{tcolorbox}
\\

\midrule

\textbf{ArXiv} &
\begin{tcolorbox}[
    colback=green!2,
    colframe=green!45,
    boxrule=0.5pt,
    arc=2mm,
    left=1.5mm,right=1.5mm,
    top=1mm,bottom=1mm,
    width=\linewidth,
    before skip=0pt,
    after skip=0pt
]
Rephrase the abstract of an article titled
\textit{``Global structure and physical interpretation of the Fonarev solution for a scalar field with exponential potential.''}

\smallskip
\textit{``We discuss the physical interpretation of a dynamical and inhomogeneous spherically symmetric solution obtained by Fonarev for a scalar field with an exponential potential. \ldots{} This demonstrates a weak point of the local definition of a black hole in terms of a trapping horizon.''}
\end{tcolorbox}
\\

\midrule

\textbf{Reddit} &
\begin{tcolorbox}[
    colback=orange!2,
    colframe=orange!55,
    boxrule=0.5pt,
    arc=2mm,
    left=1.5mm,right=1.5mm,
    top=1mm,bottom=1mm,
    width=\linewidth,
    before skip=0pt,
    after skip=0pt
]
I will ask you a question. Provide an answer with more than 200 words in \textit{a 10-year-old child’s voice}. The child has limited vocabulary and knowledge and may make grammatical or factual mistakes.

\smallskip
\textbf{Question:}
\textit{``What contingency plans did the United States have in the event that Hitler defeated the Soviet Union and/or Britain?''}
\end{tcolorbox}
\\

\midrule

\shortstack[l]{\textbf{Story}\\\textbf{Generation}} &
\begin{tcolorbox}[
    colback=purple!2,
    colframe=purple!45,
    boxrule=0.5pt,
    arc=2mm,
    left=1.5mm,right=1.5mm,
    top=1mm,bottom=1mm,
    width=\linewidth,
    before skip=0pt,
    after skip=0pt
]
\textsc{[wp]} \textit{``You are an aged witch doctor for your local tribe. You know your health is failing, so you recruit a bright young child to eventually become your successor. Today's the day that you finally show them the dark truth behind your abilities.''}
\end{tcolorbox}
\\
\bottomrule
\end{tabularx}
\caption{Example prompts from the five domains used in the human vs.\ AI-generated analysis. Colored boxes are used for illustration purposes only and do not indicate labels or categories beyond domain membership.}
\label{tab:prompt_examples}
\end{table*}

\section{Feature Taxonomy}
\label{app:features}

\begin{table*}[h]
\centering
\small
\renewcommand{\arraystretch}{1.35}
\begin{tabularx}{\textwidth}{@{} l l X l @{}}
\toprule
\textbf{\#} & \textbf{Feature} & \textbf{Definition} & \textbf{Formula} \\
\midrule
1 & \texttt{lexical\_diversity}
  & Type-token ratio: fraction of token positions occupied by unique word
    types (lowercased). Measures vocabulary reuse within the passage.
  & $|\mathcal{V}| \;/\; N$ \\
2 & \texttt{lexical\_density}
  & Fraction of tokens that are alphabetic and not NLTK English stopwords.
    Proxy for content-word density.
  & $N_{\text{content}} \;/\; N$ \\
3 & \texttt{percent\_long\_words}
  & Fraction of tokens that are alphabetic and exceed 6 characters.
  & $N_{>6} \;/\; N$ \\
4 & \texttt{entropy}
  & Word-level Shannon entropy over the empirical token frequency
    distribution of the passage. Higher values indicate more uniform vocabulary spread; lower values
    indicate a few types dominate.
  & $-\sum_{w} p(w)\log_2 p(w)$ \\
5 & \texttt{burstiness}
  & Coefficient of variation of the word frequency vector.
    Measures unevenness of word-type usage: high values indicate
    heavy-tailed distributions where a few types dominate.
  & $\sigma(f) \;/\; \mu(f)$ \\
6 & \texttt{percent\_vowels}
  & Fraction of characters (lowercased) in $\{$a, e, i, o, u$\}$.
  & $N_v \;/\; N_c$ \\
7 & \texttt{percent\_consonants}
  & Fraction of characters (lowercased) in the 21-consonant set.
  & $N_k \;/\; N_c$ \\
8 & \texttt{percent\_punctuation}
  & Fraction of characters (raw, case-sensitive) in
    \texttt{string.punctuation} (32 ASCII punctuation characters).
  & $N_p \;/\; N_c$ \\
9  & \texttt{num\_words}
   & Total token count from \texttt{word\_tokenize}, including
     punctuation tokens. Serves as a document-length feature and as the
     denominator for all ratio features above.
   & $N$ \\
10 & \texttt{num\_sentences}
   & Total sentence count from \texttt{sent\_tokenize}
   & $S$ \\
11 & \texttt{avg\_sentence\_length}
   & Mean tokens per sentence.
   & $N \;/\; S$ \\
12 & \texttt{parse\_tree\_depth}
   & Maximum number of ancestors of any token in the spaCy dependency
     parse of the full passage. Measures the deepest syntactic embedding
     present.
   & $\max_{t \in \text{doc}} |\text{ancestors}(t)|$ \\
13 & \texttt{gunning\_fog}
   & Estimates years of formal education needed to read the passage.
     Combines average sentence length with proportion of complex words
     ($\geq$3 syllables).
   & $0.4 \times (\text{ASL} + \text{PCW} \times 100)$ \\
14 & \texttt{linsear\_write}
   & Grade-level readability metric weighting polysyllabic words
     (originally developed for U.S. government writing).
   & $(N_{\text{easy}} + 3N_{\text{hard}}) \;/\; S \;/\; 2$ \\
\bottomrule
\end{tabularx}
\caption{The 14 stylometric features used in both settings.
$N$: total token count; $\mathcal{V}$: unique token types (lowercased);
$S$: sentence count; $N_c$: total character count;
$f$: word frequency vector; $\sigma$, $\mu$: standard deviation and mean.
Entropy and burstiness are both derived from the same word frequency
counter but capture distinct properties: entropy measures distributional
uniformity on a log scale; burstiness measures linear-scale frequency
dispersion. All features are computed per passage; no corpus-level statistics are used.}
\label{tab:features}
\end{table*}

We compute stylometric features that require no language model access, no external API calls, and no corpus-level statistics. All features are computed independently from the raw text of each passage using NLTK's \texttt{word\_tokenize} and \texttt{sent\_tokenize}, the spaCy \texttt{en\_core\_web\_sm} dependency parser, and the \texttt{textstat} library.

We begin with a pool of 26 candidate features spanning lexical, character-level, syntactic complexity, information-theoretic, and readability-based measures. Lexical features capture vocabulary usage patterns such as diversity and the proportion of complex words. Repetition-based features measure lexical redundancy, reflecting how frequently words or structures are reused within a passage. Readability features quantify text difficulty using established formulas that assess sentence length, word complexity, and vocabulary familiarity; examples include \textit{Flesch--Kincaid Grade Level}, \textit{Flesch Reading Ease}, and \textit{Dale--Chall}, which are among the features removed during selection due to high correlation with retained readability measures or negligible incremental predictive gain.

To arrive at a compact, non-redundant set, we apply four selection criteria sequentially, removing any feature that fails at least one: (i)~algebraic dependence on shared primitives, (ii)~pairwise Pearson correlation ($|r| > 0.9$),
(iii)~variance inflation factor ($\text{VIF} > 10$), and (iv)~negligible incremental predictive gain across all five domains. This yields a final set of 14 features, summarized in Table~\ref{tab:features}. The same 14 features are used across all experimental conditions, enabling direct comparison between global, domain-specific, and LLM-specific models.


\section{Detailed Analysis of Human vs.\ AI-Generated Detection}
\label{app:analysis_setting1}

\subsection{Classification performance and feature importance}

\begin{table*}[t]
\centering
\small
\renewcommand{\arraystretch}{1.22}

\begin{tabular}{llc p{5.2cm}}
\toprule
\textbf{Level} & \textbf{Condition} & \textbf{Acc. (\%) $\pm$ Std. (\%)} & \textbf{Top-5 Features} \\
\midrule

\rowcolor{GlobalCol}
\textbf{Global} 
& All 
& 91.74 $\pm$ 0.61 
& lexical diversity, entropy, burstiness, lexical density, word count \\
\midrule

\rowcolor{DomainCol}
\multicolumn{4}{l}{\textbf{Per-Domain Conditions}} \\
\rowcolor{DomainCol}
& ArXiv & 92.57 $\pm$ 1.29 & lexical diversity, word count, \% long words, gunning fog, \% punctuation \\
\rowcolor{DomainCol}
& Reddit & 92.67 $\pm$ 1.21 & lexical density, lexical diversity, burstiness, word count, entropy \\
\rowcolor{DomainCol}
& Story Generation & 95.83 $\pm$ 0.99 & lexical diversity, burstiness, \% consonants, \% vowels, sentence count \\
\rowcolor{DomainCol}
& WikiHow & 95.72 $\pm$ 1.01 & lexical density, entropy, lexical diversity, word count, \% vowels \\
\rowcolor{DomainCol}
& Wikipedia & 92.99 $\pm$ 1.31 & lexical diversity, entropy, burstiness, \% punctuation, \% long words \\
\midrule

\rowcolor{LLMCol}
\multicolumn{4}{l}{\textbf{Per-LLM Conditions}} \\
\rowcolor{LLMCol}
& Gemma 12B & 98.15 $\pm$ 0.32 & entropy, lexical diversity, \% long words, burstiness, \% punctuation \\
\rowcolor{LLMCol}
& Gemma 27B & 98.11 $\pm$ 0.27 & entropy, word count, lexical diversity, lexical density, \% punctuation \\
\rowcolor{LLMCol}
& GPT-OSS 20B & 98.77 $\pm$ 0.25 & lexical diversity, \% punctuation, entropy, avg.\ sent.\ length, lexical density \\
\rowcolor{LLMCol}
& GPT-OSS 120B & 79.45 $\pm$ 0.87 & burstiness, entropy, word count, sentence count, linsear write \\
\rowcolor{LLMCol}
& Llama 8B & 97.78 $\pm$ 0.31 & lexical diversity, entropy, \% long words, lexical density, \% punctuation \\
\rowcolor{LLMCol}
& Llama 70B & 95.38 $\pm$ 0.44 & entropy, word count, lexical diversity, \% consonants, burstiness \\
\rowcolor{LLMCol}
& Qwen 7B & 99.09 $\pm$ 0.21 & lexical diversity, entropy, \% long words, \% punctuation, sentence count \\
\rowcolor{LLMCol}
& Qwen 72B & 92.47 $\pm$ 0.60 & entropy, word count, burstiness, lexical diversity, lexical density \\
\bottomrule
\end{tabular}

\caption{
Accuracy summary across global, per-domain, and per-LLM evaluation settings, together with the top-5 stylometric features ranked by permutation importance.
Accuracies are reported as mean $\pm$ standard deviation over 10 runs with class-balanced subsampling.
Color shading separates the three evaluation granularities for readability.
}
\label{tab:acc_summary_full}
\end{table*}

\begin{table*}[t]
\centering
\small
\setlength{\tabcolsep}{4pt}
\begin{tabular}{lccccc}
\toprule
LLM & ArXiv & Reddit & Story Gen & WikiHow & Wikipedia \\
\midrule
Gemma 12B     & 99.55 $\pm$ 0.32 & 99.51 $\pm$ 0.35 & 99.58 $\pm$ 0.32 & 98.39 $\pm$ 0.65 & 98.43 $\pm$ 0.63 \\
Gemma 27B     & 99.45 $\pm$ 0.36 & 99.22 $\pm$ 0.47 & 99.72 $\pm$ 0.26 & 98.47 $\pm$ 0.58 & 98.55 $\pm$ 0.58 \\
GPT-OSS 20B   & 98.23 $\pm$ 0.65 & 99.74 $\pm$ 0.23 & 98.73 $\pm$ 0.55 & 99.21 $\pm$ 0.45 & 99.62 $\pm$ 0.31 \\
GPT-OSS 120B  & 84.24 $\pm$ 1.80 & 87.31 $\pm$ 1.64 & 95.68 $\pm$ 0.99 & 93.95 $\pm$ 1.13 & 89.85 $\pm$ 1.53 \\
Llama 8B      & 97.52 $\pm$ 0.75 & 99.28 $\pm$ 0.42 & 99.25 $\pm$ 0.42 & 98.49 $\pm$ 0.61 & 96.46 $\pm$ 0.95 \\
Llama 70B     & 97.32 $\pm$ 0.81 & 98.17 $\pm$ 0.67 & 97.42 $\pm$ 0.79 & 97.70 $\pm$ 0.75 & 95.29 $\pm$ 1.09 \\
Qwen 7B       & 98.80 $\pm$ 0.56 & 99.93 $\pm$ 0.14 & 99.82 $\pm$ 0.18 & 99.57 $\pm$ 0.32 & 99.15 $\pm$ 0.42 \\
Qwen 72B      & 90.41 $\pm$ 1.39 & 97.35 $\pm$ 0.79 & 98.95 $\pm$ 0.54 & 95.80 $\pm$ 0.99 & 94.75 $\pm$ 1.01 \\
\bottomrule
\end{tabular}
\caption{Pairwise accuracy ($\pm$ standard deviation over 10 runs) across domains and LLMs.}
\label{tab:pairwise_acc}
\end{table*}

\begin{figure}[t]
    \centering
    \includegraphics[width=\linewidth]{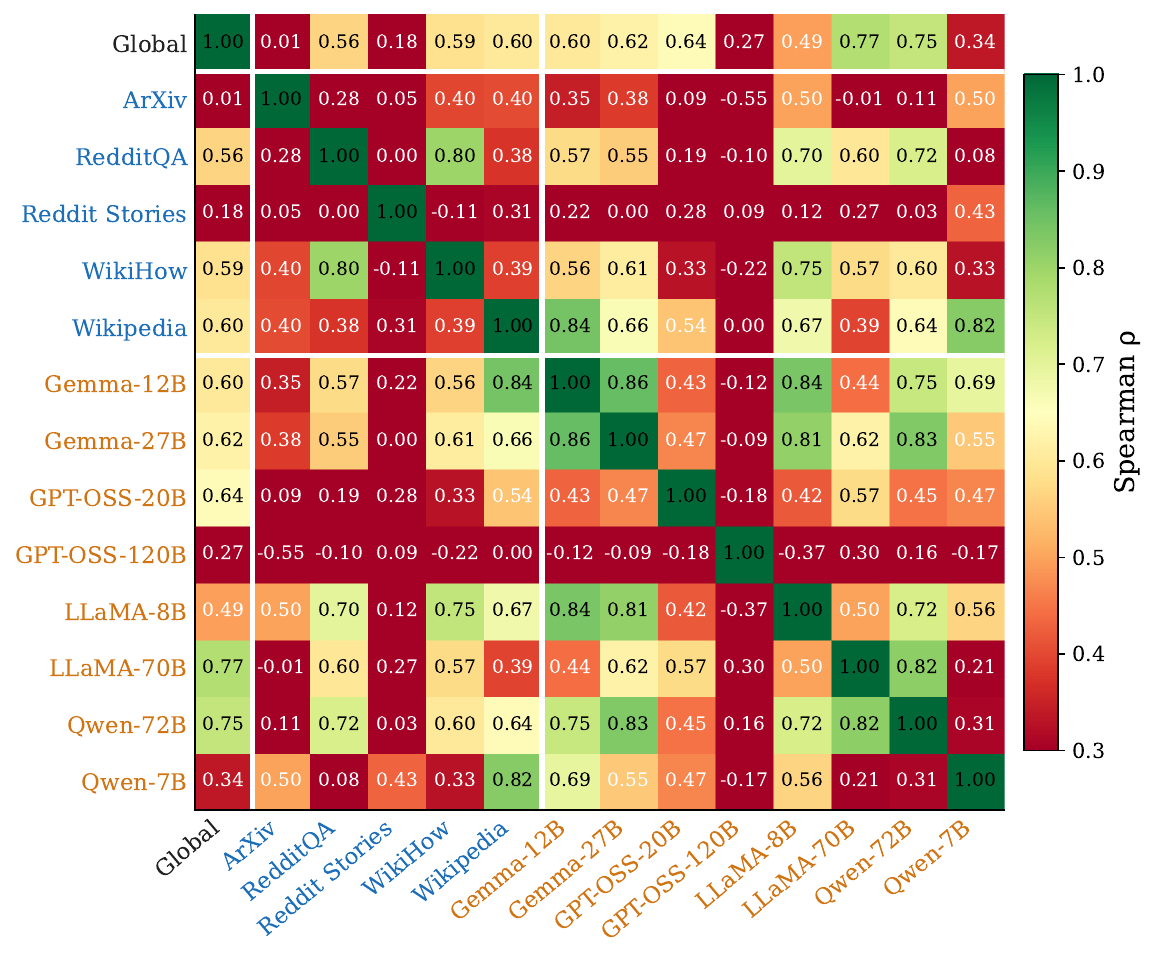}
    \caption{
    Pairwise similarity of feature-rank structures across evaluation conditions.
    Each cell shows the Spearman correlation between the 14-dimensional feature-rank vectors of two conditions, where features are ranked by permutation importance.
    }
    \label{fig:condition_similarity}
\end{figure}

\begin{figure*}[t]
    \centering
    \includegraphics[width=0.9\linewidth]{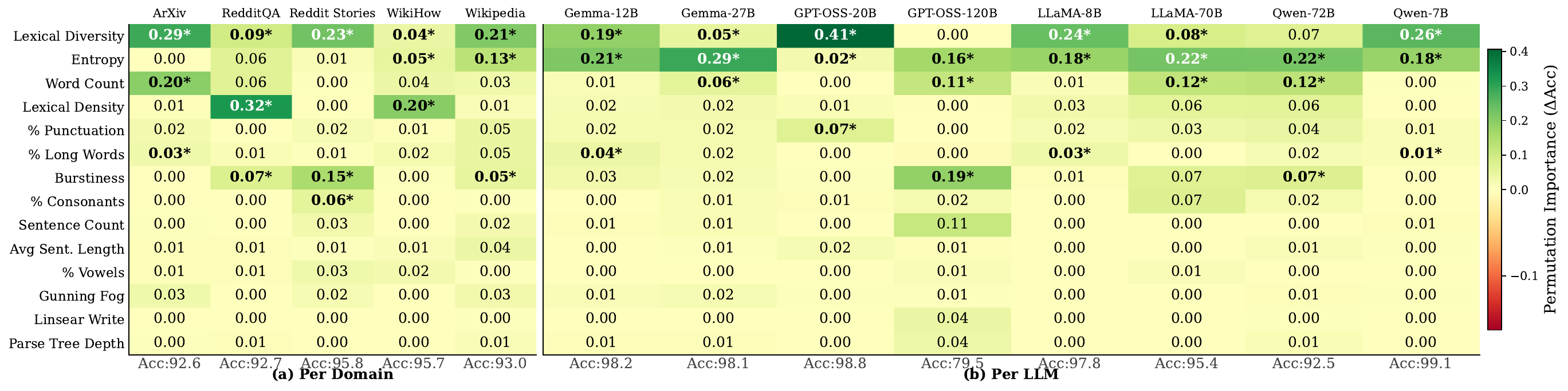}
    \caption{
    Permutation importance across domains and LLMs. Columns are condition-specific classifiers; values show held-out accuracy drop ($\Delta$Acc). Markers indicate the top-3 features per condition.
    }
    \label{fig:domain_llm}
\end{figure*}

\begin{figure*}[h]
    \centering
    \includegraphics[width=0.8\textwidth]{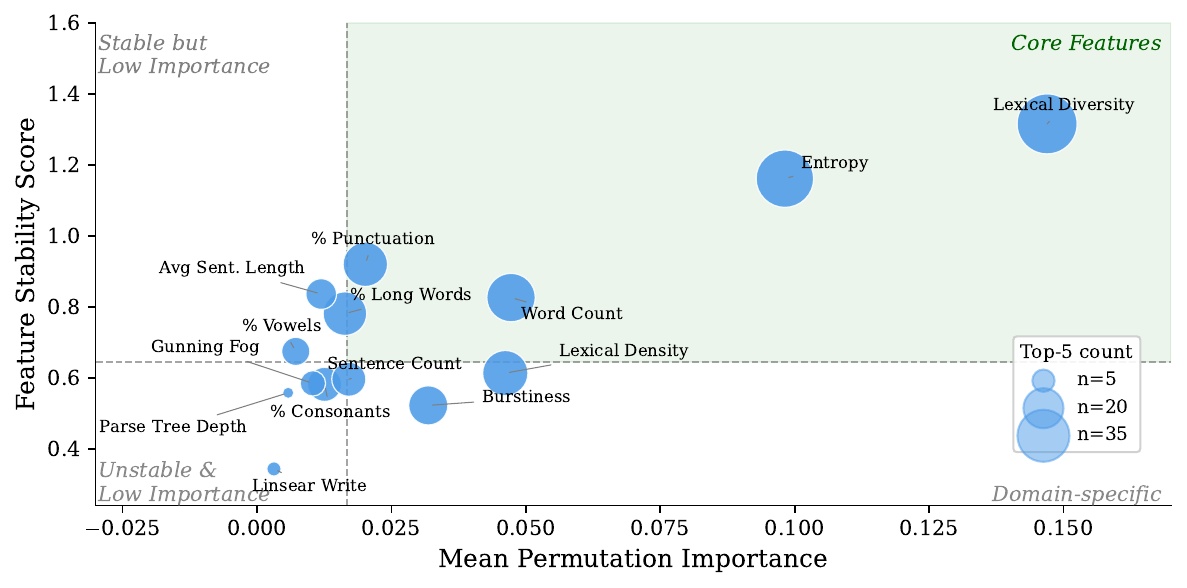}
    \caption{
    Feature robustness landscape across 54 evaluation conditions (1 global + 5 domain-specific + 8 generator-specific + 40 domain $\times$ generator pair models). x-axis: mean permutation importance across all conditions. y-axis: stability measured as 1/CV (coefficient of variation of permutation importance), where higher values indicate more consistent importance across conditions. Bubble size encodes how many of the 54 conditions each feature appeared in the top-5 most important features. Dashed lines mark the median of each axis, dividing features into four quadrants. The shaded top-right region (Core Features) identifies features that are both consistently important and stable.
    }
    \label{fig:feat_robust}
\end{figure*}


We provide detailed analysis on classification performance and feature importance across four settings: global, domain-specific, generator-specific, and pairwise (domain $\times$ generator). Overall, stylometric features provide strong discriminative power, with consistent patterns emerging across all levels.

At the \textbf{global} level, the model achieves 91.74\% accuracy ($\pm$0.61\% over 10 class-balanced runs). The top-5 features by permutation importance are lexical diversity, entropy, burstiness, lexical density, and word count. Entropy and lexical diversity dominate the ranking, sitting alone in the high-importance, high-stability region of Figure~\ref{fig:feat_robust}, while the remaining features fall into more condition-dependent regions.

When moving to \textbf{domain-specific} models, accuracies range from 92.57\% (ArXiv) to 95.83\% (Story Generation), all at or slightly above the global baseline. Three of five domains (ArXiv, Reddit, Wikipedia) improve by less than 1.5 points over the global model, indicating that within these domains the cross-LLM signal is essentially as strong as the global signal; Story Generation and WikiHow show larger improvements ($+$4 points), suggesting that restricting to a single domain genuinely makes stylistic differences more separable in those settings. Feature importance, however, becomes substantially more heterogeneous. While lexical diversity remains in the top-5 for four of five domains, the supporting features vary: ArXiv and Wikipedia favor length and readability features (word count, \% long words, gunning fog, \% punctuation), Reddit and WikiHow are dominated by lexical density (the top feature in both, with ArXiv-like length features secondary), and Story Generation emphasizes burstiness and character composition (\% consonants, \% vowels). This indicates that domain introduces structured variation in how stylometric signals manifest, even as a small core feature set remains stable.

In contrast, \textbf{LLM-specific} models exhibit a much wider spread in performance, with accuracies ranging from 79.45\% (GPT-OSS 120B) to 99.09\% (Qwen 7B) --- a 19.6-point gap. Either entropy or lexical diversity is the top feature for seven of eight LLMs, but GPT-OSS 120B emerges as a clear outlier on both axes: it places burstiness first, followed by entropy, word count, sentence count, and linsear write, and its accuracy is more than 10 points below the next-lowest LLM (Qwen 72B, 92.47\%). The remaining seven LLMs combine entropy with lexical diversity and varying secondary features --- typically word count, lexical density, \% long words, or \% punctuation --- and all achieve accuracy above 95\%.

The \textbf{pairwise} (domain $\times$ LLM) analysis (Table~\ref{tab:pairwise_acc}) provides the most fine-grained view. Across all 40 pairs the mean accuracy is 97.24\%, but systematic differences emerge. Pairs involving the Qwen and Gemma families consistently achieve near-perfect accuracy ($>$98\% across most domains), while GPT-OSS 120B remains the most challenging generator at every domain (mean pairwise accuracy 90.21\%), with its lowest accuracy on ArXiv (84.24\%) and Reddit (87.31\%). Domain effects also stack with LLM effects: Story Generation pairs are uniformly easy (mean 98.64\% across all eight LLMs), while ArXiv pairs span the widest range (84.24--99.55\%), reflecting genuine difficulty in distinguishing AI from human writing in the academic-abstract register. Local difficulty spots also appear outside the GPT-OSS 120B column --- Qwen 72B $\times$ ArXiv (90.41\%) and Llama 70B $\times$ Wikipedia (95.29\%) --- suggesting that some difficult pairs involve domain-specific writing styles (academic abstracts, encyclopedic prose) that even strong LLMs do not perfectly imitate.

Taken together, these results reveal a consistent structure across all levels. First, stylometric classification remains highly effective, with accuracy at or above 92\% in every domain and seven of eight per-LLM settings. Second, a stable core of universal features --- lexical diversity and entropy --- appears in the top-5 of nearly every condition, supporting the cross-condition feature-rank stability documented in Section~5.3. Third, additional features are context-dependent, varying systematically with domain (length and readability features for ArXiv and Wikipedia, lexical density for Reddit and WikiHow, character composition for Story Generation) and with LLM (most strikingly, the burstiness-dominated signature of GPT-OSS 120B).

\subsection{Condition-level similarity analysis}
\label{app:condition_similarity}

To evaluate whether different conditions share similar feature structures, we compute pairwise Spearman correlations between the 14-dimensional feature-rank vectors of each condition, where ranks are assigned according to permutation importance. Figure~\ref{fig:condition_similarity} visualizes the resulting condition-similarity matrix with hierarchical clustering based on $1-\rho$.

Overall agreement is moderate but heterogeneous, with a median pairwise correlation of $\rho = 0.43$ and values ranging from $-0.55$ to $0.86$. Most per-LLM conditions form a relatively tight central cluster, indicating that many generators share similar feature-ordering structures. In particular, Gemma-12B, Gemma-27B, Llama-8B, Llama-70B, Qwen-72B, and the global condition exhibit broadly consistent rankings centered around entropy and lexical diversity.

GPT-OSS-120B is the primary outlier. Its mean correlation with the remaining 13 conditions is negative ($\bar{\rho} = -0.075$), and it achieves the lowest pairwise agreement in the matrix ($\rho = -0.55$ with ArXiv). It is the only condition whose feature ordering is on average negatively correlated with the broader consensus. This behavior is consistent with the distinct importance profile identified in Section~\ref{sec:cross_condition_variability}, where burstiness outranks entropy only for GPT-OSS-120B.

Per-domain conditions exhibit greater heterogeneity overall. Story Generation and ArXiv emerge as the most distinctive domains, with mean correlations substantially below the remaining conditions. Their feature structures rely more heavily on condition-specific signals such as burstiness and word count, pushing them away from the entropy/lexical-diversity-dominated consensus shared by most other settings. In contrast, Wikipedia, Reddit, and WikiHow remain substantially closer to the global structure.

These condition-level differences mirror the per-condition variability observed in Figure~\ref{fig:domain_llm}: conditions with the most distinctive top-ranked features also exhibit the lowest agreement with the broader cross-condition consensus. Together, the correlation analysis and feature-level stability analysis support the same decomposition --- a small transferable core of stable features combined with a larger condition-specific shell.

\subsection{Leave-one-LLM-out stability}
\label{app:loo_llm}

To evaluate whether the observed feature rankings depend heavily on any single generator family, we perform leave-one-LLM-out (LOO-LLM) experiments. For each fold, one LLM is excluded during training and evaluation is performed using the remaining seven generators. We then compare the resulting feature rankings using both Spearman correlation and top-$k$ overlap statistics.

Across the eight LOO-LLM folds, coefficient rankings remain highly consistent. The mean pairwise Spearman correlation between feature-rank vectors is $\rho = 0.92$, while the mean top-5 Jaccard overlap is 0.84. In practice, removing any single generator typically preserves four of the five most important features.

Entropy and lexical diversity remain consistently top-ranked across nearly all folds, indicating that their importance does not depend on any particular model family. In contrast, lower-ranked features such as readability metrics and sentence-level statistics exhibit substantially greater variability, with occasional rank reversals across folds.

GPT-OSS-120B again emerges as the most influential held-out condition. Excluding GPT-OSS-120B increases average agreement across the remaining folds, consistent with its distinctive feature-ordering structure identified in Section~\ref{app:condition_similarity}. Nevertheless, even when GPT-OSS-120B is excluded, entropy and lexical diversity remain dominant, indicating that the broader cross-condition structure is not driven solely by this outlier.

Overall, these experiments suggest that the primary feature rankings remain stable under generator-level perturbation and are not tied to any single LLM family.

\subsection{Leave-one-domain-out stability}
\label{app:loo_domain}

We next evaluate leave-one-domain-out (LOO-domain) stability by excluding one domain at a time during training and comparing the resulting feature rankings across the remaining domains.

Stability remains substantial but is consistently lower than in the LOO-LLM setting. Across the five folds, the mean pairwise Spearman correlation is $\rho = 0.83$, and the mean top-5 Jaccard overlap is 0.80. This reduction is consistent with the stronger domain-level heterogeneity identified in Section~\ref{sec:cross_condition_variability}, where dominant feature rankings vary more across domains than across generators.

ArXiv and Story Generation produce the largest ranking shifts when excluded, reflecting their more distinctive stylometric structures. In particular, excluding ArXiv substantially reduces the importance of length and readability features, while excluding Story Generation reduces the prominence of burstiness and character-composition features.

Despite these shifts, entropy and lexical diversity remain among the top-ranked features in every fold. Their stability across both domain- and generator-level perturbations further supports their role as the most transferable stylometric features in the analysis.

\subsection{Sign consistency analysis}
\label{app:sign_consistency}

Beyond rank stability, we also analyze whether feature directions remain consistent across conditions. For each feature and each leave-one-out fold, we record the sign of the standardized logistic regression coefficient. Across all 13 leave-one-out folds (8 LOO-LLM and 5 LOO-domain), 10 of the 14 features retain the same coefficient sign in every fold. The most stable features include lexical diversity, entropy, and burstiness, all of which consistently receive positive coefficients. These results indicate that AI-generated text tends to exhibit higher values along these dimensions across the evaluated settings. Features that occasionally flip sign are generally low-ranked by importance, including several readability and sentence-level metrics. Their instability is therefore concentrated among weaker, condition-specific signals rather than the dominant cross-condition features. Taken together, the sign-consistency results reinforce the broader stability findings throughout Section~5: the primary stylometric structure remains largely stable across domains and generators, while instability is concentrated among lower-importance features.

\subsection{Length Controls}
\label{app:length_controls}

A potential concern is that the two most stable features in our main analysis, lexical diversity and entropy, may partly reflect length artifacts rather than stylometric differences. Type-token ratio is sensitive to document length, entropy depends on the empirical word-frequency distribution, and generated outputs are subject to a maximum token budget. We therefore conduct three complementary controls to test whether the main feature-stability result is driven by document length or truncation effects.

\paragraph{Removing explicit length features.}
We first remove the two features that directly encode document length, \texttt{num\_words} and \texttt{num\_sentences}, and refit the global logistic regression model using the remaining 12 features under the same 10-run balanced protocol as the main analysis. Accuracy decreases only slightly, from 91.74\% to 90.98\% (95\% CI [90.57, 91.38]). Four of the top five features remain unchanged: lexical diversity, entropy, burstiness, and lexical density. The only change is that \texttt{num\_words} is replaced by percent punctuation. This indicates that the dominant lexical-diversity and entropy signals are not reducible to explicit document-length features.

\paragraph{Length-stratified analysis.}
We next divide the corpus into four quartiles by \texttt{num\_words} and refit the classifier separately within each length range. If lexical diversity or entropy were primarily length proxies, their importance should diminish when between-text length variation is minimized. Instead, all four quartile-restricted classifiers remain accurate: 92.96\% in Q1, 93.65\% in Q2, 90.22\% in Q3, and 95.27\% in Q4. Lexical diversity is the top-ranked feature in every quartile, with permutation importance $\Delta$Acc = 0.449, 0.355, 0.216, and 0.362 respectively. Notably, lexical diversity is most important in the Q1 and Q2 quartile. Entropy remains most prominent in the shorter texts and becomes less dominant in the longest quartile.

\paragraph{Long-output truncation check.}
Finally, we test whether very long generated outputs near the maximum token budget drive the global signal. For each generator, we remove the top 10\% longest texts using generator-specific word-count cutoffs, and apply the same trimming procedure to human texts. After trimming, the classifier's accuracy remains essentially unchanged, decreasing from 91.74\% to 91.64\% (95\% CI [91.24, 92.04]). The top-ranked features are also stable: lexical diversity, entropy, burstiness, and lexical density retain the first four positions, with percent punctuation replacing \texttt{num\_words} in the fifth position.

Together, these controls suggest that the main stability result is not driven solely by explicit length features, within-length variation, or long-output truncation. Lexical diversity remains consistently informative even under strict length stratification, while entropy remains part of the global stable feature set, though its within-bin contribution is strongest for shorter texts.

\subsection{Sampling-Temperature Ablation}
\label{sec:temperature-ablation}

Our main generation experiments use temperature 0.7. Because sampling temperature directly affects token-distribution variability, we test whether the importance of entropy and lexical diversity depends on this choice. We sample 1,000 prompts, with 200 prompts from each of
the five domains, and generate one document per prompt using Llama-3.1-8B at temperatures 0.3, 0.7, and 1.0. Prompts, top-$p$, maximum output length, and the feature-extraction and classification pipelines are held fixed across conditions. Because this ablation uses a fixed 1,000-prompt subset, its absolute accuracy is not directly comparable to the full per-LLM result in Table~\ref{tab:acc_summary}; we use this experiment to compare relative changes across temperatures under an otherwise fixed setup.

\begin{table*}[t]
    \centering
    \small
    \resizebox{\linewidth}{!}{%
    \begin{tabular}{lccc}
        \toprule
        Temperature & Accuracy (\%) &  Top-5 Features\\
        \midrule
        0.3 & 82.2 & lexical diversity, entropy, \% consonants, word count, \% vowels \\
        0.7 & 80.7 &  lexical diversity, word count, \% consonants, entropy, \% vowels\\
        1.0 & 79.5 &  lexical diversity, word count, \% consonants, burstiness, entropy\\
        \bottomrule
    \end{tabular}}
    \caption{Human-written vs.\ AI-generated classification across
    sampling temperatures. Feature ranks are based on permutation
    importance.}
    \label{tab:temperature-ablation}
\end{table*}

As shown in Table~\ref{tab:temperature-ablation}, classification accuracy decreases modestly as temperature increases. The ordering of feature importance remains substantially stable with pairwise Spearman correlations between the complete 14 feature rankings ranging from $\rho=0.86$ to $0.93$, with a mean of $0.89$.

Lexical diversity is the highest-ranked feature under every temperature. Entropy remains among the five most informative features, although its relative contribution decreases at higher temperatures, where features such as document length and burstiness become more prominent. We therefore do not claim that every feature has identical importance under all decoding settings. Rather, the results show that the core generation-associated signal, particularly lexical diversity together with entropy, remains informative across a substantial range of sampling temperatures.

\subsection{Proprietary-Model Generation}
\label{sec:proprietary-generation}

The main experiments use open-weight generators so that generation parameters can be controlled consistently and the resulting corpus can be reproduced. To examine whether the observed stylometric separation extends to proprietary systems, we conduct an additional targeted analysis using Gemini-3-Flash and GPT-5.4-mini.

For each model, we generate 2,500 documents using 500 prompts from each of the five domains. We apply the same feature-extraction pipeline and class-balanced Human vs.\ AI-Gen classification protocol used in the main analysis.

\begin{table*}[t]
    \centering
    \small
    \begin{tabular}{llp{9.2cm}}
        \toprule
        Model &  Accuracy (\%) & Top-5 Features \\
        \midrule
        Gemini-3-Flash
        & 82.1
        & lexical diversity, \% consonants, burstiness,
          \% vowels, \% long words \\

        GPT-5.4-mini
        & 78.6
        & \% consonants, Gunning Fog, \% vowels,
          lexical diversity, burstiness \\
        \bottomrule
    \end{tabular}
    \caption{Human-written vs.\ AI-generated classification for the
    proprietary-generator subset. Features are ordered by permutation
    importance.}
    \label{tab:proprietary-generation}
\end{table*}

Table~\ref{tab:proprietary-generation} shows that outputs from both proprietary generators remain distinguishable from human-written text using the same stylometric feature space. Lexical diversity remains highly informative, ranking first for Gemini-3-Flash and fourth for GPT-5.4-mini. The remaining feature rankings are more generator-dependent. Gemini-3-Flash retains a pattern centered on
lexical diversity and burstiness, whereas GPT-5.4-mini is separated more strongly by character-composition and readability features, including the percentages of consonants and vowels and Gunning Fog. These results extend the overall stylometric detectability pattern beyond the open-weight models in the main benchmark. At the same time, they reinforce the distinction between a limited transferable core and a larger set of generator-dependent secondary signals.


\section{Implementation Detail}
\label{app:implementation-detail}
 
\paragraph{Training and evaluation split.}
For the AI-edit vs.\ AI-gen and Human vs.\ AI-edit tasks, we hold out 20 of the 1,000 seed texts for classifier training and tuning. Each seed yields up to 279 edited variants, producing 5,580 edited texts, from which we sample 2,000 balanced training instances and 1,000 evaluation instances per task. The remaining 980 seeds (273,420 edited texts) are used exclusively for analysis. Splitting at the seed level ensures no edited variant of a training seed appears in the analysis set.
 
\paragraph{EditLens} EditLens is the state-of-the-art editing-aware detector with demonstrated generalization across domains and editing conditions. We use two pretrained checkpoints directly, \texttt{editlens\_roberta-large} and \texttt{editlens\_Llama-3.2-3B} (QLoRA 4-bit), without further training, and adopt their classification formulation rather than their regression variant. Inference is run with a maximum sequence length of 512 tokens for RoBERTa and 1,024 tokens for Llama, in bf16 precision. Each text receives a single continuous score between 0 and 1, where higher values indicate greater AI involvement. Since EditLens does not produce task-specific predictions, we determine a binary threshold for each task via grid search on the test set, maximizing macro F1 on the corresponding pair. We note that the cosine-distance thresholds EditLens uses to define its training buckets are label thresholds on the cosine scale, distinct from the decision thresholds we select here on the model's output score. For the ternary analysis in Section~\ref{app:ternary}, we additionally partition the score axis into three zones using two thresholds ($t_1$, $t_2$), determined by maximizing 3-class macro F1 on the test set (1,000 human + 1,000 edited + 1,000 AI-generated). For \texttt{editlens\_Llama}, this yields $t_1=0.16$ and $t_2=0.63$: scores below 0.16 are classified as human, 0.16--0.63 as edited, and above 0.63 as AI-generated.
 
\section{Classifier Validation}
\label{app:validation}
 
Before applying classifiers to the full analysis set, we verify that all three methods produce meaningful predictions on a test set (1,000 samples per class). Table~\ref{tab:validation} reports the accuracy and AUC for each task. All three methods achieve $\geq$92.8\% accuracy in distinguishing human-written text from AI-generated text. When separating edited text from generated text, LR (93.2\%) outperforms EditLens (65.8--77.6\%). In detecting editing against human-written text, EditLens (93.1\%) outperforms LR (69.8\%).
 
\begin{table}[ht]
\centering
\small
\resizebox{\columnwidth}{!}{%
\begin{tabular}{llccc}
\toprule
Task & Metric & LR & EL-R & EL-L \\
\midrule
Human vs. AI-gen & Acc & 93.4 & 92.8 & 98.7 \\
              & AUC & 0.97 & 0.97 & 1.00 \\
\midrule
AI-edit vs. AI-gen & Acc & 93.2 & 65.8 & 77.6 \\
              & AUC & 0.98 & 0.71 & 0.85 \\
\midrule
Human vs. AI-edit  & Acc & 69.8 & 87.8 & 93.1 \\
              & AUC & 0.80 & 0.94 & 0.97 \\
\bottomrule
\end{tabular}%
}
\caption{GEN:Edit\_test performance. EditLens thresholds selected via grid search (Section~\ref{app:implementation-detail}).}
\label{tab:validation}
\end{table}
 
\section{Additional Results by Editing Condition}
\label{app:full_results}
 
Table~\ref{tab:app_273k_full} reports the correct classification rate on the 273{,}420-text editing corpus by edit ratio, under each method's task-specific optimal threshold.
 
\begin{table}[ht]
\centering
\footnotesize
\resizebox{\columnwidth}{!}{%
\begin{tabular}{lr cccc}
\toprule
Edit ratio & $n$ & LR & EL-R & EL-L & LR$+$EL-L \\
\midrule
\multicolumn{6}{l}{\textit{Human vs.\ AI-gen}} \\
\midrule
$\le$ 20\%& 6{,}416   & 8.5  & 54.4 & 14.6 & 4.2  \\
20--40\%  & 25{,}789  & 17.3 & 84.6 & 64.1 & 20.2 \\
40--60\%  & 96{,}223  & 26.9 & 90.7 & 85.8 & 41.4 \\
60--80\%  & 141{,}718 & 32.4 & 91.8 & 94.3 & 60.3 \\
$\ge$ 80\% & 3{,}274   & 26.5 & 89.4 & 92.0 & 58.9 \\
\midrule
\multicolumn{6}{l}{\textit{AI-edit vs.\ AI-gen}} \\
\midrule
$\le$ 20\%& 6{,}416   & 92.7 & 94.2 & 99.8 & 98.8 \\
20--40\%  & 25{,}789  & 91.7 & 70.7 & 92.6 & 95.7 \\
40--60\%  & 96{,}223  & 93.0 & 50.2 & 70.7 & 94.3 \\
60--80\%  & 141{,}718 & 93.6 & 42.9 & 46.2 & 91.3 \\
$\ge$ 80\% & 3{,}274   & 91.8 & 31.1 & 40.5 & 85.8 \\
\midrule
\multicolumn{6}{l}{\textit{Human vs.\ AI-edit}} \\
\midrule
$\le$ 20\%& 6{,}416   & 68.0 & 75.0 & 46.8 & 67.8 \\
20--40\%  & 25{,}789  & 80.5 & 92.5 & 88.0 & 89.4 \\
40--60\%  & 96{,}223  & 87.0 & 94.1 & 96.1 & 96.9 \\
60--80\%  & 141{,}718 & 89.2 & 94.4 & 98.9 & 98.3 \\
$\ge$ 80\% & 3{,}274   & 81.3 & 92.2 & 97.5 & 96.2 \\
\bottomrule
\end{tabular}}
\caption{Correct classification rate (\%) on the 273{,}420-text editing corpus by actual edit ratio, under each method's task-specific optimal threshold.}
\label{tab:app_273k_full}
\end{table}
 
\subsection{Threshold Sensitivity}
\label{app:threshold}
 
\Cref{tab:auc} reports AUC, a threshold-free metric, while Tables~\ref{tab:app_optimal} and \ref{tab:app_tpr} evaluate performance under tuned thresholds and fixed FPRs. Results are consistent across all settings. LR’s optimal threshold remains close to $0.5$ ($0.47$ and $0.51$), suggesting minimal calibration needs. The same complementary pattern holds across all evaluation settings: LR achieves a lower TPR on Human vs.\ AI-edit detection but a higher TPR on AI-edit vs.\ AI-gen detection, whereas EL-L shows the opposite trend.
 
\begin{table}[ht]
\centering
\small
\resizebox{\columnwidth}{!}{%
\begin{tabular}{lccc}
\toprule
Method &
\shortstack[c]{Human\\vs.\ AI-gen} &
\shortstack[c]{AI-edit\\vs.\ AI-gen} &
\shortstack[c]{Human\\vs.\ AI-edit} \\
\midrule
LR        & 0.47 / 93.6 & 0.51 / 93.2 & 0.36 / 72.5 \\
EL-R      & 0.12 / 93.0 & 0.34 / 68.3 & 0.07 / 89.1 \\
EL-L      & 0.28 / 98.9 & 0.53 / 79.8 & 0.14 / 93.3 \\
\midrule
LR$+$EL-L & 0.44 / 97.5 & 0.53 / 95.3 & 0.32 / 86.1 \\
\bottomrule
\end{tabular}
}
\caption{Optimal decision threshold and corresponding accuracy (\%).}
\label{tab:app_optimal}
\end{table}

\begin{table}[ht]
\centering
\small
\setlength{\tabcolsep}{11pt}
\resizebox{\columnwidth}{!}{%
\begin{tabular}{lccc}
\toprule
& \multicolumn{3}{c}{Predicted} \\
\cmidrule(lr){2-4}
True & Human & AI-edit & AI-gen \\
\midrule
Human   & 0.745 & 0.191 & 0.064 \\
AI-edit & 0.319 & 0.626 & 0.055 \\
AI-gen  & 0.050 & 0.058 & 0.892 \\
\bottomrule
\end{tabular}
}
\caption{Confusion matrix for three-class classification.}
\label{tab:three_class_confusion}
\end{table}

\begin{table*}[ht]
\centering
\small
\resizebox{\textwidth}{!}{%
\begin{tabular}{lccccccccc}
\toprule
 & \multicolumn{3}{c}{FPR $=1\%$} & \multicolumn{3}{c}{FPR $=5\%$} & \multicolumn{3}{c}{FPR $=10\%$} \\
\cmidrule(lr){2-4}\cmidrule(lr){5-7}\cmidrule(lr){8-10}
Method & \makecell{Human\\ vs.\\ AI-gen} & \makecell{AI-edit\\ vs.\\ AI-gen} & \makecell{Human\\ vs.\\ AI-edit} & \makecell{Human\\ vs.\\ AI-gen} & \makecell{AI-edit\\ vs.\\ AI-gen} & \makecell{Human\\ vs.\\ AI-edit} & \makecell{Human\\ vs.\\ AI-gen} & \makecell{AI-edit\\ vs.\\ AI-gen} & \makecell{Human\\ vs.\\ AI-edit} \\
\midrule
LR        & 80.1 & 78.5 & 17.5 & 91.3 & 91.0 & 33.8 & 94.8 & 94.6 & 47.0 \\
EL-R      & 81.9 & 10.6 & 60.6 & 90.8 & 23.8 & 79.8 & 93.3 & 32.2 & 87.9 \\
EL-L      & 98.7 & 6.9 & 76.0 & 99.3 & 29.6 & 91.2 & 99.6 & 54.2 & 93.3 \\
\midrule
LR$+$EL-L & 94.9 & 86.7 & 56.0 & 98.1 & 95.2 & 70.7 & 99.7 & 96.3 & 81.9 \\
\bottomrule
\end{tabular}%
}
\caption{Detection rate (TPR, \%) at fixed false-positive rates on the test set.}
\label{tab:app_tpr}
\end{table*}

\subsection{Three-Class Classification}
\label{app:three_class}

To examine how the three categories are positioned in a single feature space, we train a multinomial logistic regression over Human, AI-edit, and AI-gen, using the same 14 features and the same train/test split as the pairwise experiments in Section~\ref{sec:phase2-results}. The classifier achieves 0.754 accuracy, 0.753 macro-F1, and 0.910 macro-AUC (one-vs-rest). Per-class F1 is 0.887 for AI-gen, 0.705 for Human, and 0.668 for AI-edit.

\autoref{tab:three_class_confusion} shows the confusion matrix. AI-generated text remains clearly separable (recall 89.2\%), consistent with the pairwise results. AI-edited text is positioned asymmetrically toward the human side: its misclassifications overwhelmingly go to Human (31.9\%) rather than to AI-gen (5.5\%), consistent with Human vs.\ AI-edit being the hardest pairwise task.

\subsection{EditLens Ternary Analysis}
\label{app:ternary}
 
Using thresholds $t_1 = 0.16$, $t_2 = 0.63$ for EL-L (determined by maximizing 3-class macro F1 on the test set), we partition the 273k edited texts into three classes.
 
\begin{table}[ht]
\centering
\small
\resizebox{\columnwidth}{!}{%
\begin{tabular}{lccc}
\toprule
Condition & Human & AI-edit & AI-gen \\
\midrule
\textbf{Overall} & 5.2 & 67.8 & 27.0 \\
\midrule
\multicolumn{4}{l}{\textit{By category}} \\
Grammar      & 41.7 & 57.3 & 1.1 \\
Concision    & 3.9 & 75.4 & 20.7 \\
Paraphrasing & 4.7 & 75.2 & 20.1 \\
Fluency      & 4.3 & 71.6 & 24.1 \\
Tone         & 4.4 & 67.2 & 28.4 \\
Adding detail & 3.9 & 65.0 & 31.1 \\
Structure    & 3.5 & 61.9 & 34.6 \\
Clarity      & 2.9 & 59.6 & 37.5 \\
\midrule
\multicolumn{4}{l}{\textit{By edit degree}} \\
0--20\%      & 58.0 & 42.0 & 0.1 \\
20--40\%     & 14.4 & 82.8 & 2.8 \\
40--60\%     & 4.8 & 77.7 & 17.5 \\
60--80\%     & 1.4 & 59.9 & 38.6 \\
80--100\%    & 3.0 & 50.0 & 47.0 \\
\midrule
\multicolumn{4}{l}{\textit{By editing model}} \\
Llama        & 5.0 & 54.6 & 40.4 \\
Qwen         & 5.8 & 65.2 & 29.1 \\
GPT-OSS      & 4.8 & 83.7 & 11.5 \\
\bottomrule
\end{tabular}}
\caption{EditLens-Llama ternary zone distribution (\%) by condition. Human zone: score $< 0.16$; Edited zone: $0.16$--$0.63$; AI-Generated zone: $> 0.63$.}
\label{tab:app_ternary}
\end{table}
 
\subsection{Cohen's $d$ by Editing Category and Domain}
\label{app:cohens_d_breakdown}
 
Tables~\ref{tab:app_cohens_category} and \ref{tab:app_cohens_domain} report Cohen's $d$ between human-written seeds and their edited counterparts, broken down by editing category and domain respectively.
 
\begin{table*}[ht]
\centering
\begin{tabular}{lcccccccc}
\toprule
Feature & Gram. & Flu. & Tone & Para. & Clar. & Add. & Conc. & Struct. \\
\midrule
Lex.\ density     & -0.01 & 0.38 & 0.17 & 0.40 & 0.45 & 0.30 & 0.41 & 0.39 \\
Lex.\ diversity   & -0.30 & 0.64 & 0.49 & 0.62 & 0.66 & 0.07 & 0.80 & 0.49 \\
\% long words     & 0.13 & 0.53 & 0.31 & 0.62 & 0.71 & 0.38 & 0.39 & 0.62 \\
Entropy           & 0.03 & -0.08 & 0.12 & -0.10 & -0.17 & 0.47 & -0.61 & -0.06 \\
Burstiness        & 0.08 & -0.06 & 0.07 & -0.10 & -0.06 & 0.74 & -0.41 & 0.11 \\
Gunning fog       & 0.01 & 0.30 & 0.20 & 0.29 & 0.44 & 0.42 & -0.02 & 0.38 \\
Linsear write     & -0.30 & -0.05 & -0.12 & -0.14 & -0.00 & 0.12 & -0.49 & -0.05 \\
Word count        & 0.03 & -0.41 & -0.27 & -0.42 & -0.47 & 0.19 & -0.70 & -0.38 \\
Sentence count    & 0.14 & -0.37 & -0.27 & -0.33 & -0.42 & -0.07 & -0.50 & -0.38 \\
Avg.\ sent.\ len  & -0.14 & -0.00 & 0.01 & -0.10 & -0.01 & 0.26 & -0.33 & -0.02 \\
Parse tree depth  & -0.08 & -0.25 & -0.17 & -0.25 & -0.28 & 0.14 & -0.61 & -0.21 \\
\% vowels         & 0.17 & 0.33 & 0.29 & 0.37 & 0.41 & 0.24 & 0.05 & 0.39 \\
\% consonants     & 0.18 & 0.51 & 0.33 & 0.46 & 0.47 & 0.56 & 0.25 & 0.53 \\
\% punctuation    & 0.12 & -0.13 & -0.06 & -0.13 & -0.08 & -0.19 & 0.20 & -0.09 \\
\bottomrule
\end{tabular}
\caption{Cohen's $d$ by editing category.}
\label{tab:app_cohens_category}
\end{table*}
 
\begin{table*}[t]
\centering
\begin{tabular}{lccccc}
\toprule
Feature & ArXiv & Reddit & Story & WikiHow & Wikipedia \\
\midrule
Lex.\ density     & 0.19 & 1.04 & 0.45 & 0.95 & -0.56 \\
Lex.\ diversity   & 0.39 & 0.62 & 0.35 & 0.75 & 0.52 \\
\% long words     & 0.31 & 1.13 & 0.94 & 0.83 & 0.05 \\
Entropy           & 0.69 & -0.29 & 0.01 & -0.03 & -0.31 \\
Burstiness        & 0.29 & 0.08 & 0.25 & -0.17 & 0.03 \\
Gunning fog       & 0.26 & 0.50 & 0.67 & 0.62 & 0.02 \\
Linsear write     & -0.24 & -0.25 & 0.17 & -0.00 & -0.19 \\
Word count        & 0.50 & -0.43 & -0.15 & -0.53 & -0.47 \\
Sentence count    & 0.49 & -0.34 & -0.45 & -0.48 & -0.43 \\
Avg.\ sent.\ len  & -0.03 & -0.19 & 0.57 & -0.11 & -0.09 \\
Parse tree depth  & -0.15 & -0.26 & -0.02 & -0.19 & -0.37 \\
\% vowels         & 0.01 & 0.34 & 0.95 & 0.20 & 0.11 \\
\% consonants     & 0.15 & 0.47 & 0.87 & 0.65 & 0.14 \\
\% punctuation    & 0.01 & -0.28 & -0.37 & -0.03 & 0.43 \\
\bottomrule
\end{tabular}
\caption{Cohen's $d$ by domain.}
\label{tab:app_cohens_domain}
\end{table*}
 
\subsection{Token-Level Patterns in Edited Text}
\label{app:token_signals}
 
Certain transition phrases are heavily overrepresented in edited text relative to human text: ``notably'' appears 27.1$\times$ more frequently, ``consequently'' 20.6$\times$, and ``additionally'' 6.1$\times$. Conversely, phrases like ``hence'' (0.17$\times$) and ``it is important to'' (0.18$\times$) are underrepresented. These patterns are directly accessible to EditLens through raw text input but are not captured by any of the 14 stylometric features.
 
\begin{figure*}[t]
    \centering
    \includegraphics[width=\textwidth]{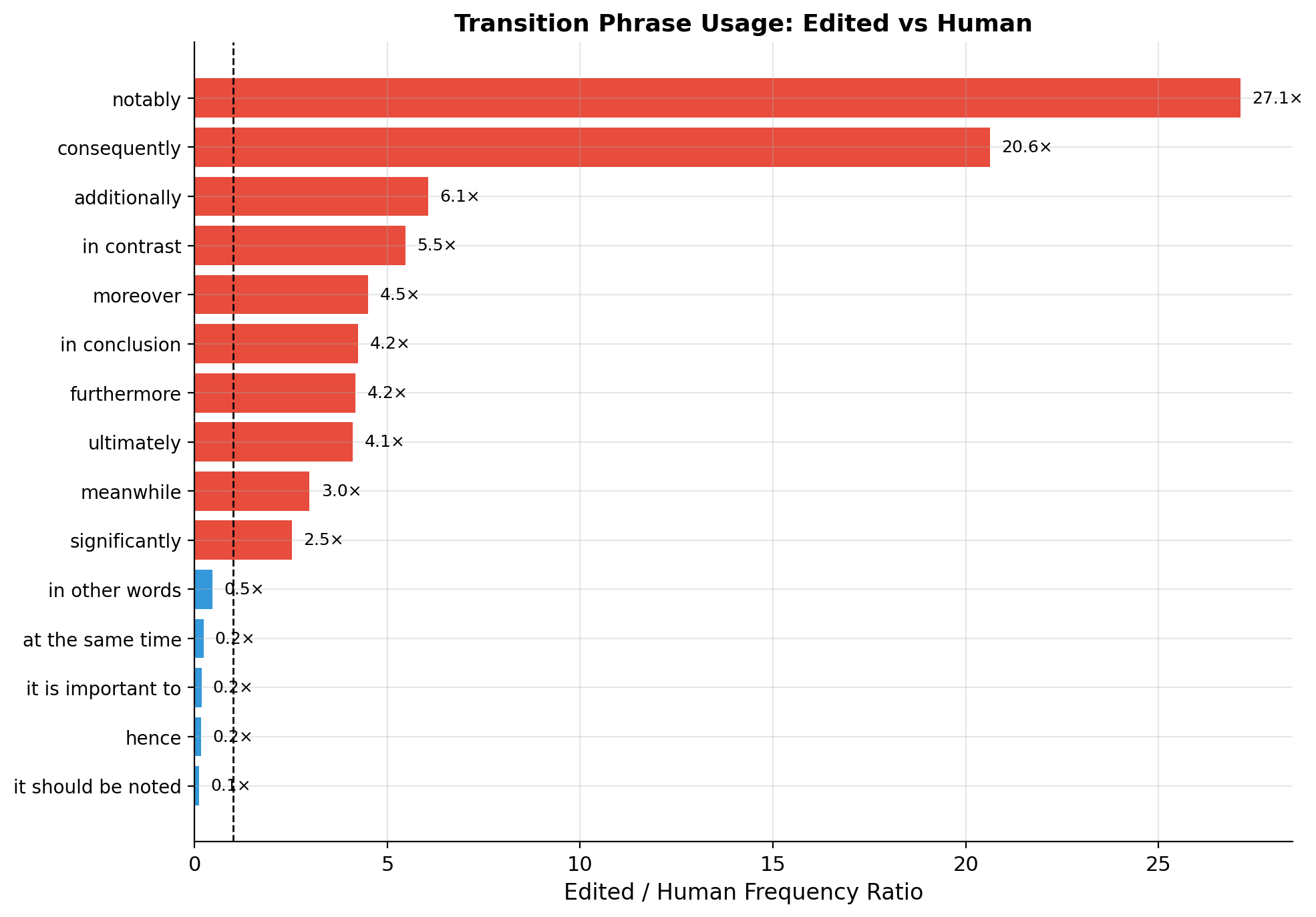}
    \caption{Transition phrase frequency in AI-edited text relative to human writing.}
\label{fig:ngram_ratio}
\end{figure*}
 
\subsection{Editing Prompts}
\label{app:editing_prompts}

\autoref{tab:editing_prompts} shows the full editing prompts.
 
\begin{table*}[t]
\centering
\small
\resizebox{\linewidth}{!}{%
\begin{tabular}{@{}lll@{}}
\toprule
\textbf{Category} & \textbf{Prompt} & \textbf{Source} \\
\midrule
\multirow{6}{*}{Tone \& Style}
 & Lighten the tone and add a touch of humor & Gemini 2.5 Pro \\
 & Adjust the tone to be more friendly & Claude Sonnet 4 \\
 & Increase the emotional stakes & Claude Sonnet 4 \\
 & Make this sound more formal and school-like & ChatGPT 4o \\
 & Make this nicer but keep the main point & ChatGPT 4o \\
 & Change this so people from other countries can get it too & ChatGPT 4o \\
\midrule
\multirow{6}{*}{Adding Detail}
 & Make this longer with more evidence & Human \\
 & Expand this text with more specific examples and details & Claude Sonnet 4 \\
 & Show rather than tell by adding scene-setting details & Claude Sonnet 4 \\
 & Include sounds, smells, textures, and other sensory elements & Claude Sonnet 4 \\
 & Add a short story to make this more interesting & ChatGPT 4o \\
 & Show how this works in real life & ChatGPT 4o \\
\midrule
\multirow{5}{*}{Concision}
 & Simplify this text & Human \\
 & Make this more specific & Human \\
 & Reduce wordiness while amplifying impact & Claude Sonnet 4 \\
 & Say this in fewer words without losing meaning & ChatGPT 4o \\
 & Edit this to be punchier and more direct & ChatGPT 4o \\
\midrule
\multirow{4}{*}{Fluency \& Flow}
 & Make this text flow more naturally & Claude Sonnet 4 \\
 & Improve how the sentences sound together & ChatGPT 4o \\
 & Refine the pacing and cadence of this paragraph & ChatGPT 4o \\
 & Write a better and more interesting beginning & ChatGPT 4o \\
\midrule
\multirow{4}{*}{Clarity \& Precision}
 & Write this in a way that my teacher would get it & Human \\
 & Can you make my paper more persuasive? & Human \\
 & Make the main idea clearer & ChatGPT 4o \\
 & Improve this to make it stronger and clearer & ChatGPT 4o \\
\midrule
\multirow{3}{*}{Paraphrasing}
 & Paraphrase this & Human \\
 & Rewrite this in different words while keeping the same meaning & Claude Sonnet 4 \\
 & Reword this to improve clarity while keeping the meaning & ChatGPT 4o \\
\midrule
\multirow{2}{*}{Structure \& Org.}
 & Make sure there's a beginning, middle, and end & ChatGPT 4o \\
 & Group related ideas and use transitions to improve structure & ChatGPT 4o \\
\midrule
Grammar \& Mech. & Can you fix any spelling, grammar, or punctuation issues? & Human \\
\bottomrule
\end{tabular}%
}
\caption{Editing prompts used in AI-Edited Text Analysis, drawn from the EditLens test split~\citep{thai2026editlens}. Prompts are grouped by category.}
\label{tab:editing_prompts}
\end{table*}

\section{Model Licenses}
\label{app:model_licenses}

Table~\ref{tab:model_licenses} summarizes the open-weight LLMs used for AI text generation, including their Hugging Face identifiers and licenses or usage terms. We report the license associated with each checkpoint at the time of data generation.

\begin{table*}[t]
\centering
\setlength{\tabcolsep}{4pt}
\renewcommand{\arraystretch}{1.12}
\begin{tabular}{lll}
\toprule
\textbf{Model} & \textbf{Hugging Face Identifier} & \textbf{License / Terms} \\
\midrule
\texttt{Qwen-2.5-7B} 
& \texttt{Qwen/Qwen2.5-7B} 
& Apache 2.0 \\

\texttt{Qwen-2.5-72B} 
& \texttt{Qwen/Qwen2.5-72B} 
& Qwen License \\

\texttt{Gemma-3-12B} 
& \texttt{google/gemma-3-12b-it} 
& Gemma Terms of Use \\

\texttt{Gemma-3-27B} 
& \texttt{google/gemma-3-27b-it} 
& Gemma Terms of Use \\

\texttt{Llama-3.1-8B} 
& \texttt{meta-llama/Llama-3.1-8B-Instruct} 
& Llama 3.1 Community License \\

\texttt{Llama-3.3-70B} 
& \texttt{meta-llama/Llama-3.3-70B-Instruct} 
& Llama 3.3 Community License \\

\texttt{GPT-oss-20B} 
& \texttt{openai/gpt-oss-20b} 
& Apache 2.0 \\

\texttt{GPT-oss-120B} 
& \texttt{openai/gpt-oss-120b} 
& Apache 2.0 \\
\bottomrule
\end{tabular}
\caption{
Hugging Face identifiers and licenses or usage terms for the open-weight LLMs.
}
\label{tab:model_licenses}
\end{table*}

\end{document}